\documentclass[twoside, a4paper, 12pt, bibliography=totoc]{scrartcl}
\usepackage[shortcuts]{extdash}
\usepackage{graphicx}
\usepackage{xcolor}
\usepackage{pifont}
\usepackage[a4paper]{anysize}
\usepackage[round,colon]{natbib}
\usepackage{times}
\usepackage[hyphens]{url}
\usepackage[german,english]{babel}
\usepackage{moreverb}
\usepackage[fleqn]{amsmath}
\usepackage{amsfonts}
\usepackage{amssymb}
\usepackage{amsthm}
\usepackage{thmtools}
\usepackage{graphicx}
\usepackage{environ}
\usepackage{bm}
\usepackage[utf8]{inputenc}
\usepackage[T1]{fontenc}
\usepackage{hyperref}
\usepackage{breakurl}
\usepackage{nameref}
\marginsize{3.0cm}{3.0cm}{2.0cm}{2.0cm}
\usepackage[right]{showlabels}

\usepackage{mathtools}
\usepackage{caption}
\usepackage{subcaption}

\definecolor{blue4}{HTML}{00008E}
\definecolor{red2}{HTML}{CE0000}

\declaretheoremstyle[
  headfont=\color{blue4}\normalfont\bfseries,
  bodyfont=\color{blue4}\normalfont\itshape,
]{colored}

\newcommand{\trarxiv}[2]{#2}
\begin{document}
\pagenumbering{gobble}

% changing line labels in the moreverb package (listing env.)
\def\listinglabel#1{\llap{\tiny\ttfamily\the#1}\hskip\listingoffset\relax}

\newcommand{\trtitle}{Improved Convergence Speed of Fully Symmetric
  Learning Rules for Principal Component Analysis}
\newcommand{\tryear}{2020}

%===========================================================================
\trarxiv{%
}{%
  \titlehead{%
    \includegraphics[width=8cm]{Faculty-of-Technology_vektorisiert.eps}}
}{%
  \titlehead{\hspace*{1cm}}
}
\subject{\hspace*{1cm}}
\title{%
  \vspace*{-3cm}
  \trtitle%
}
\trarxiv{%
  \author{\textsf{Ralf Möller}}
}{%
  \author{%
    Ralf Möller\\
    Computer Engineering Group, Faculty of Technology\\
    Bielefeld University, Bielefeld, Germany\\
    \url{www.ti.uni-bielefeld.de}
  }
}
\trarxiv{%
  \date{\normalsize\textsf{\tryear\\[5mm]version of \today}}
}{%
  \date{\hspace*{1cm}}
}
\maketitle
%===========================================================================

%\input /home/moeller/tex/mathcmds-1-4.tex
%===========================================================================
%
% mathcmds.tex --
%
%
% Ralf Moeller <moeller@mpipf-muenchen.mpg.de>
%
%   Copyright (C) 2002
%   Cognitive Robotics Group
%   Max Planck Institute for Psychological Research
%   Munich
%
% 1.0 / 10. Jul 02 (rm)
% - from scratch
% 1.1 / 24. Oct 04 (rm)
% - pmat, rank, sgn
% 1.2 / 25. Oct 04 (rm)
% - pihalf
% 1.3 /  8. Feb 05 (rm)
% - fixed bug in \matN
% 1.4 / 16. Jan 06 (rm)
% - new: grad
% 1.5 / 22. Jan 06 (rm)
% - gradv, df, matGamma, matBeta
% 1.6 / 25. Apr 06 (rm)
% - ddfs, ddfsq
% 1.7 /  7. Mar 07 (rm)
% - vecmu, vecnu, cov
% 1.8 / 12. Feb 08 (rm)
% - matPhi
% 1.9 / 20. May 08 (rm)
% - atantwo operator
% 1.10 /  7. Oct 08 (rm)
% - vecunull, vecvnull
% 1.11 /  9. Oct 08 (rm)
% - matUnull, matVnull, matSnull
% 1.12 / 26. Feb 09 (rm)
% - asin operator
% - eps shorthand for varepsilon
% 1.13 /  2. Mar 09 (rm)
% - acos operator
% 1.14 /  4. Aug 09 (rm)
% - atan operator
% 1.15 /  3. Aug 10 (rm)
% - vecone
% 1.16 /  2. Nov 15 (rm)
% - \bf -> \mathbf
% 1.17 / 19. Jan 17 (rm)
% - mathcmds1-2!
% - vecxdot, corrections to other dot commands
% 1.18 / 20. May 19 (rm)
% - matAnull
% - \bar -> \bar{}
% - \rm -> \textrm
% 1.19 / 27. May 19 (rm)
% - \betanull
%
%===========================================================================

% sums etc.
\newcommand{\summe}[2]{\sum\limits_{#1}^{#2}}
\newcommand{\produkt}[2]{\prod\limits_{#1}^{#2}}
\newcommand{\sumin}{\summe{i=1}{n}}
\newcommand{\sumjn}{\summe{j=1}{n}}
\newcommand{\sumkn}{\summe{k=1}{n}}
\newcommand{\sumim}{\summe{i=1}{m}}
\newcommand{\sumjm}{\summe{j=1}{m}}
\newcommand{\prodin}{\produkt{i=1}{n}}
\newcommand{\prodim}{\produkt{i=1}{m}}
\newcommand{\prodjm}{\produkt{j=1}{m}}
% matrices
\newcommand{\matA}{\mathbf{A}}
\newcommand{\matAnull}{\bar{\mathbf{A}}}
\newcommand{\matB}{\mathbf{B}}
\newcommand{\matBnull}{\bar{\mathbf{B}}}
\newcommand{\matC}{\mathbf{C}}
\newcommand{\matD}{\mathbf{D}}
\newcommand{\matDnull}{\bar{\mathbf{D}}}
\newcommand{\matE}{\mathbf{E}}
\newcommand{\matF}{\mathbf{F}}
\newcommand{\matG}{\mathbf{G}}
\newcommand{\matH}{\mathbf{H}}
\newcommand{\matI}{\mathbf{I}}
\newcommand{\matJ}{\mathbf{J}}
\newcommand{\matK}{\mathbf{K}}
\newcommand{\matL}{\mathbf{L}}
\newcommand{\matM}{\mathbf{M}}
\newcommand{\matMnull}{\bar{\mathbf{M}}}
\newcommand{\matN}{\mathbf{N}}
\newcommand{\matO}{\mathbf{O}}
\newcommand{\matP}{\mathbf{P}}
\newcommand{\matPhi}{\boldsymbol{\Phi}}
\newcommand{\matQ}{\mathbf{Q}}
\newcommand{\matR}{\mathbf{R}}
\newcommand{\matS}{\mathbf{S}}
\newcommand{\matSnull}{\bar{\mathbf{S}}}
\newcommand{\matT}{\mathbf{T}}
\newcommand{\matTheta}{\boldsymbol{\Theta}}
\newcommand{\matU}{\mathbf{U}}
\newcommand{\matUnull}{\bar{\mathbf{U}}}
\newcommand{\matV}{\mathbf{V}}
\newcommand{\matVnull}{\bar{\mathbf{V}}}
\newcommand{\matW}{\mathbf{W}}
\newcommand{\matWnull}{\bar{\mathbf{W}}}
\newcommand{\matX}{\mathbf{X}}
\newcommand{\matXnull}{\bar{\mathbf{X}}}
\newcommand{\matY}{\mathbf{Y}}
\newcommand{\matZ}{\mathbf{Z}}
% special
\newcommand{\matWdot}{{\dot{\mathbf{W}}}}
\newcommand{\matWe}{{\mathbf{W}^e}}
\newcommand{\matBeta}{\mathbf{B}}
\newcommand{\matDelta}{{\mathbf\Delta}}
\newcommand{\matGamma}{{\mathbf\Gamma}}
\newcommand{\matLambda}{{\mathbf\Lambda}}
\newcommand{\matOmega}{{\mathbf\Omega}}
\newcommand{\matXi}{{\mathbf\Xi}}
\newcommand{\matLambdanull}{\bar{\mathbf{\Lambda}}}
\newcommand{\matNull}{\mathbf{0}}
% vectors
\newcommand{\veca}{\mathbf{a}}
\newcommand{\vecalpha}{\boldsymbol{\alpha}}
\newcommand{\vecb}{\mathbf{b}}
\newcommand{\vecc}{\mathbf{c}}
\newcommand{\vecd}{\mathbf{d}}
\newcommand{\vece}{\mathbf{e}}
\newcommand{\veceta}{\boldsymbol{\eta}}
\newcommand{\vecf}{\mathbf{f}}
\newcommand{\vecg}{\mathbf{g}}
\newcommand{\vech}{\mathbf{h}}
\newcommand{\veci}{\mathbf{i}}
\newcommand{\vecj}{\mathbf{j}}
\newcommand{\veck}{\mathbf{k}}
\newcommand{\vecl}{\mathbf{l}}
\newcommand{\vecm}{\mathbf{m}}
\newcommand{\vecmnull}{\bar{\mathbf{m}}}
\newcommand{\vecmu}{\boldsymbol{\mu}}
\newcommand{\vecn}{\mathbf{n}}
\newcommand{\vecnu}{\boldsymbol{\nu}}
\newcommand{\veco}{\mathbf{o}}
\newcommand{\vecp}{\mathbf{p}}
\newcommand{\vecphi}{\boldsymbol{\varphi}}
\newcommand{\vecq}{\mathbf{q}}
\newcommand{\vecr}{\mathbf{r}}
\newcommand{\vecs}{\mathbf{s}}
\newcommand{\vect}{\mathbf{t}}
\newcommand{\vecu}{\mathbf{u}}
\newcommand{\vecunull}{\bar{\mathbf{u}}}
\newcommand{\vecv}{\mathbf{v}}
\newcommand{\vecvnull}{\bar{\mathbf{v}}}
\newcommand{\vecx}{\mathbf{x}}
\newcommand{\vecxdot}{\dot{\mathbf{x}}}
\newcommand{\vecxxi}{\boldsymbol{\xi}}
\newcommand{\vecy}{\mathbf{y}}
\newcommand{\vecz}{\mathbf{z}}
\newcommand{\veczeta}{\boldsymbol{\zeta}}
\newcommand{\vecw}{\mathbf{w}}
\newcommand{\vecwe}{{\mathbf{w}^e}}
\newcommand{\vecwei}{{\mathbf{w}^e_i}}
\newcommand{\vecwek}{{\mathbf{w}^e_k}}
\newcommand{\vecwdot}{\dot{\mathbf{w}}}
\newcommand{\vecwnull}{\bar{\mathbf{w}}}
\newcommand{\vecdwdt}{\frac{d\vecw}{dt}}
\newcommand{\vecnull}{\mathbf{0}}
\newcommand{\vecone}{\mathbf{1}}
\newcommand{\vecvdot}{\dot{\mathbf{v}}}
\newcommand{\vecomega}{\boldsymbol{\omega}}
% norms
\newcommand{\norm}[1]{\|#1\|}
\newcommand{\normF}[1]{\|#1\|_F}
\newcommand{\normm}{\norm{\vecm}}
\newcommand{\normw}{\norm{\vecw}}
\newcommand{\normx}{\norm{\vecx}}
\newcommand{\xp}[1]{\langle #1\rangle}
\newcommand{\xpvecm}{\xp{\vecm}}
\newcommand{\normxpm}{\norm{\xpvecm}}
% other
\newcommand{\betanull}{\bar{\beta}}
\newcommand{\lambdanull}{\bar{\lambda}}
\newcommand{\Wnull}{\bar{W}}
\newcommand{\half}{\frac{1}{2}}
\newcommand{\third}{\frac{1}{3}}
\newcommand{\quarter}{\frac{1}{4}}
\newcommand{\order}[1]{{\cal O}(#1)}
\newcommand{\ddt}{\frac{d}{dt}}
\newcommand{\vecstk}[1]{\left(\begin{array}{c}#1\end{array}\right)}
\newcommand{\blkstk}[1]{\left(\begin{array}{c|c}#1\end{array}\right)}
\newcommand{\pmat}[1]{\begin{pmatrix}#1\end{pmatrix}}
%
% mathops (like log or limsup)
%\def\tr{\mathop{\textrm{tr}}\nolimits}
%\def\diag{\mathop{\textrm{diag}}\nolimits}
%\newcommand{\Diag}[2]{\mathop{\textrm{diag}}\limits_{#1}^{#2}}
%\def\cov{\mathop{\textrm{cov}}\nolimits}
%\def\var{\mathop{\textrm{var}}\nolimits}
% new version:  5. Jul 19 (rm)
\newcommand{\tr}{\operatorname{tr}\nolimits}
\newcommand{\dg}{\operatorname{dg}\nolimits}
\newcommand{\diag}{\operatorname{diag}\nolimits}
\newcommand{\Diag}[2]{\operatorname*{diag}\limits_{#1}^{#2}}
\newcommand{\cov}{\operatorname{cov}\nolimits}
\newcommand{\var}{\operatorname{var}\nolimits}
\newcommand{\sign}{\operatorname{sign}\nolimits}
\newcommand{\blkdiag}{\operatorname{blkdiag}\nolimits}
\newcommand{\Blkdiag}[2]{\operatorname*{blkdiag}\limits_{#1}^{#2}}
\newcommand{\mean}{\operatorname{mean}\nolimits}

\newcommand{\rank}{\operatorname{rank}}
\newcommand{\sgn}{\operatorname{sgn}}
\newcommand{\atantwo}{\operatorname{atan2}}
\newcommand{\asin}{\operatorname{asin}}
\newcommand{\acos}{\operatorname{acos}}
\newcommand{\atan}{\operatorname{atan}}
\newcommand{\adj}{\operatorname{adj}}

% partial derivative
\newcommand{\ddf}[2]{\frac{\partial #1}{\partial #2}}
\newcommand{\df}[1]{\ddf{}{#1}}
% ... second order
\newcommand{\ddfs}[3]{\frac{\partial^2 #1}{\partial #2\partial #3}}
\newcommand{\ddfsq}[2]{\frac{\partial^2 #1}{\partial {#2}^2}}
\newcommand{\pihalf}{\frac{\pi}{2}}

% \grad
\newcommand{\grad}{\nabla}
\newcommand{\gradv}{\boldsymbol{\nabla}}

% \eps
\newcommand{\eps}{\varepsilon}

%\newcommand{\mattW}{\tilde{\matW}}
%\newcommand{\vectw}{\tilde{\vecw}}
%\newcommand{\mattU}{\tilde{\matU}}
%\newcommand{\vectu}{\tilde{\vecu}}
%\newcommand{\mattV}{\tilde{\matV}}
%\newcommand{\vectv}{\tilde{\vecv}}

%===========================================================================

\trarxiv{%
}{ %
  \renewcommand{\showlabelsetlabel}[1]{}
}

%===========================================================================

\trarxiv{%
  \newcommand{\lemmaref}[1]{\textbf{\color{red2}Lemma~#1}}
  \newcommand{\lemmasep}{\vspace*{10mm}}
}{%
  \newcommand{\lemmaref}[1]{\textbf{Lemma~#1}}
  \newcommand{\lemmasep}[1]{}
}

%===========================================================================

\trarxiv{%
  \newcommand{\todo}[1]{\textbf{\color{orange}$\bullet$~TODO: #1}\\}
  \newcommand{\todocomment}[1]{\textbf{\color{orange}(TODO: #1)}}
}{%
  \newcommand{\todo}[1]{}
  \newcommand{\todocomment}[1]{}
}

%===========================================================================
%\vspace*{-1cm}
\trarxiv{\newpage}{\vspace*{-1cm}}

\begin{abstract}
\trarxiv{\noindent\sloppy \textbf{Abstract}\\[0.5cm]}{}
\noindent%
Fully symmetric learning rules for principal component analysis can be
derived from a novel objective function suggested in our previous
work. We observed that these learning rules suffer from slow
convergence for covariance matrices where some principal eigenvalues
are close to each other. Here we describe a modified objective
function with an additional term which mitigates this convergence
problem. We show that the learning rule derived from the modified
objective function inherits all fixed points from the original
learning rule (but may introduce additional ones). Also the stability
of the inherited fixed points remains unchanged. Only the steepness of
the objective function is increased in some directions. Simulations
confirm that the convergence speed can be noticeably improved,
depending on the weight factor of the additional term.

\trarxiv{%
  \vspace*{1cm}
  Please cite as: Ralf Möller. {\em \trtitle}. Technical Report, Computer
  Engineering, Faculty of Technology, Bielefeld University, \tryear,
  version of \today, \url{www.ti.uni-bielefeld.de}.
}{%
}
\end{abstract}

%===========================================================================

%\thispagestyle{empty}
\newpage
%\thispagestyle{empty}
%\begin{footnotesize}
\tableofcontents
%\end{footnotesize}
%\thispagestyle{empty}
\newpage
\pagenumbering{arabic}

%===========================================================================

\newcommand{\exteq}[1]{\textbf{(#1)}}

%===========================================================================

\sloppypar
% split eqnarray over multiple pages
% https://github.com/davidar/blog/wiki/2008-07-26-splitting-eqnarray-over-multiple-pages
\allowdisplaybreaks

\newpage

%############################################################################
%############################################################################
%############################################################################

\section{Introduction}
%=====================

In our previous work \cite[]{own_Moeller20a}, we derived several fully
symmetric learning rules\footnote{In a fully symmetric learning rule,
  all units see the same input and perform exactly the same
  computations. Earlier symmetric learning rules required a distinct
  weight factor in each unit to ensure convergence to the principal
  eigenvectors and not just to the principal subspace.} for principal
component analysis (PCA), starting from a novel objective function (in
this paper referred to as ``original'' objective function). We
analyzed the fixed points of these learning rules and (indirectly via
the objective function) their stability. We could show that the
learning rules have stable, desired fixed points in the eigenvectors
of the covariance matrix, but exhibit additional undesired fixed
points; however, the latter are unstable. Preliminary simulations
confirmed that the learning rules converge towards the desired fixed
points, but also revealed a disadvantage: If some principal
eigenvalues of the covariance matrix are close to each other, the
learning rules operate close the undesired fixed points which
noticeably slows down convergence.

In this continuation of our work, we introduce an additional term into
our objective function which mitigates the convergence problem. We
derive a learning rule from this modified objective
function.\footnote{Our analysis focuses on the simplest (``short'')
  learning rule from our previous work, since our simulations show
  that the more complex (``long'') learning rules differ only
  marginally in their behavior, probably since terms coincide in the
  vicinity of the Stiefel manifold of the eigenvector estimates; see
  appendix \ref{app_near_st}.}  We determine the fixed points of the
learning rule and show that the modified learning rule shares the
fixed points of the original one, but may introduce additional fixed
points. Using the same indirect method as in our previous work, we
study the stability at the shared fixed points and show that it is
unchanged compared to the original objective function. Simulations
confirm both the theoretical results and the improved convergence
speed of the novel learning rule.

We recapitulate the notation in section \ref{sec_notation} and our
Lagrange-multiplier approach in section \ref{sec_lagrange}. The
original objective function and the corresponding (``short'') learning
rule are recapitulated in \ref{sec_objfct_orig} together with insights
on the fixed-point structure which motivate the modifications
introduced here. Section \ref{sec_objfct_mod_1} introduces the
modified objective function from which we derive a (``short'')
learning rule in section \ref{sec_lr_mod}. The fixed points of this
modified learning rule are analyzed in section \ref{sec_M2S_fp}. The
stability of the fixed points is analyzed indirectly from the modified
objective function in section \ref{sec_stability_mod}. Simulations are
presented in section \ref{sec_simulations}. The report ends with a
discussion (section \ref{sec_discussion}) and conclusions (section
\ref{sec_conclusion}).

%############################################################################
%############################################################################
%############################################################################

\section{Notation}\label{sec_notation}
%=================

We use the same notation as in our previous work
\cite[]{own_Moeller20a}. Table \ref{tab_matrices} shows the names of
widely used matrices. Column vector $i$ of a matrix $\matX$ is written
as $\vecx_i$. Fixed-point variables are marked by a bar
(e.g. $\matWnull$). Sometimes, matrix and vector sizes are indicated
by suffixes; for vectors and symmetric matrices, only one suffix is
provided.

Table \ref{tab_operators} shows the operators used.

\begin{table}[tp]
\begin{center}
\caption{Notation: matrices}
\label{tab_matrices}
\begin{tabular}{lll}
  $\matC$ & $n \times n$
  & covariance matrix\\
  $\matV$ & $n \times n$
  & matrix of eigenvectors $\vecv_i$ (columns) of $\matC$\\
  $\matLambda$ & $n \times n$
  & diagonal matrix of eigenvalues $\lambda_i$ (distinct, descending)
  of $\matC$\\
  $\matW$ & $n \times m$
  & matrix of principal eigenvector estimates $\vecw_i$ (columns) of $\matC$\\
  $\matA$ & $n \times m$
  & projection of $\matW$ onto the eigenvectors $\matV$\\
  $\matQ$ & $n \times n$
  & orthogonal matrix: $\matQ^T \matQ = \matQ \matQ^T = \matI_n$\\
  $\matB$ & $m \times m$
  & matrix used to form the Lagrange multipliers\\
  $\matI_n$ & $n \times n$
  & unit matrix\\
  $\matNull_{n,m}$ & $n \times m$
  & null matrix\\
\end{tabular}
\end{center}
\end{table}

\begin{table}[tp]
\begin{center}
\caption{Notation: operators}
\label{tab_operators}
\begin{tabular}{ll}
  $\delta_{ij}$
  & Kronecker's delta\\
  $\dg\{\matX\}$
  & diagonal matrix with diagonal elements from $\matX$\\
  $\diag_{i=1}^{n}\{x_i\}$
  & diagonal matrix with $n$ diagonal elements $x_i$\\
  $\blkdiag_{l=1}^{k}\{\matX_l\}$
  & block-diagonal matrix with $k$ blocks $\matX_l$\\
  $\|\matX\|^2_F = \tr\{\matX^T\matX\}$
  & squared Frobenius norm of $\matX$
\end{tabular}
\end{center}
\end{table}

References to equations and lemmata from our previous work
\cite[]{own_Moeller20a} are printed in bold font.

%############################################################################
%############################################################################
%############################################################################

\section{Lagrange-Multiplier Approach}\label{sec_lagrange}
%======================================

We use the Lagrange-multiplier approach from our previous work
\cite[]{own_Moeller20a}. For a given objective function $J$, we write
the extended objective function $J^*$ as
\begin{align}
  J^*(\matB,\matW)
  =
  J(\matW)
  +
  C(\matB,\matW)
\end{align}
where $C$ is the constraint term which includes the matrix $\matB$
(elements $\beta_{jk}$) which forms the Lagrange multipliers. We use
the same symmetric construction for the Lagrange multipliers as in
equation \exteq{39}, here with $\Omega_j = 1$ (i.e. operating on a
Stiefel manifold):
\begin{align}
  C(\matB,\matW)
  =
  \half \summe{j=1}{m}\summe{k=1}{m}
  \half (\beta_{jk}+\beta_{kj}) \left(\vecw_j^T\vecw_k-\delta_{jk}\right).
\end{align}

\section{Original Objective Function}\label{sec_objfct_orig}
%====================================

The original ``novel'' objective function from equation \exteq{23} is
\begin{align}\label{eq_objfct_orig}
  J(\matW)
  =
  \quarter \summe{j=1}{m} \left(\vecw_j^T\matC\vecw_j\right)^2.
\end{align}
We are interested in the local maxima of this function. From
(\ref{eq_objfct_orig}) we derived a fully symmetric learning rule
``N2S'', either from our ``short'' form derivation \exteq{450} or from
the canonical metric on the Stiefel manifold \exteq{486}:
\begin{align}\label{eq_N2S}
  \tau \matWdot = \matC\matW\matD - \matW\matD\matW^T\matC\matW
\end{align}
where $\tau$ is a time constant and
\begin{align}
  \matD = \Diag{j=1}{m}\{\vecw_j^T\matC\vecw_j\}.
\end{align}
The fixed-point structure of this equation is relatively complex. If
all diagonal elements of $\matDnull$ are pairwise different, we obtain
the special solution \exteq{270}
\begin{align}\label{eq_N2S_fp_special}
\matWnull = \matV \matP \pmat{\matI_m\\ \matNull}
\end{align}
where $\matP$ is an arbitrary $n\times n$ permutation matrix. If some
diagonal elements of $\matDnull$ may coincide, we obtain the general
solution for the fixed points
\begin{align}\label{eq_N2S_fp_general}
\matWnull = \matV \matP \pmat{\matU^{*T} \matP^*\\ \matNull}.
\end{align}
Here $\matU^*$ is an orthogonal block-diagonal matrix (where the size
of each block depends on the number of identical diagonal elements in
$\matDnull$), and $\matP^*$ another permutation matrix (which is
chosen such that identical diagonal elements in $\matDnull$ are
contiguous in a rearranged matrix $\matDnull^*$, see \exteq{249}).

To motivate our modified objective function below, we look at the term
$\matWnull^T \matC \matWnull$. From equations \exteq{282} and
\exteq{284} we know that, in the fixed points, we have
\begin{align}\label{eq_N2S_WTCW}
  \matWnull^T \matC \matWnull
  &=
  \matP^{*T}
  \Blkdiag{l=1}{k}\{\matU^{*'}_l\hat{\matLambda}^*_l\matU^{*'T}_l\}
  \matP^*
\end{align}
under the constraint \exteq{295}
\begin{align}\label{eq_N2S_constraint}
  \dg\{\matU^{*'}_l\hat{\matLambda}^*_l\matU^{*'T}_l\}
  =
  \overline{d}^{*'}_l \matI_l.
\end{align}
Each diagonal matrix $\hat{\matLambda}^*_l$ is a block of the
upper-left $m \times m$ part of a permuted version of the eigenvalue
matrix $\matLambda$.

It is not clear which matrices $\matU_l^{*'}$ fulfill constraint
(\ref{eq_N2S_constraint}). However, we can say that if the constraint
is fulfilled and $\matU^{*'}_l$ is not of size $1 \times 1$, the
matrix $\matU^{*'}_l\hat{\matLambda}^*_l\matU^{*'T}_l$ cannot be
diagonal \exteq{Lemma~5}. Therefore the block-diagonal matrix in
(\ref{eq_N2S_WTCW}) has non-zero off-diagonal elements. A permutation
transformation of a matrix as in (\ref{eq_N2S_WTCW}) permutes the
positions of the diagonal elements (see \exteq{516}), which entails
that off-diagonal elements remain at off-diagonal positions. Therefore
$\matWnull^T \matC \matWnull$ from (\ref{eq_N2S_WTCW}) has non-zero
off-diagonal elements if we are at an {\em undesired} fixed point.

In contrast, if we look at the {\em desired} fixed points from
(\ref{eq_N2S_fp_special}), we see that $\matWnull^T \matC \matWnull$
is diagonal:\footnote{Note that $\hat{\matLambda}^*$ in this
  derivation may differ from the one in (\ref{eq_N2S_WTCW}) and
  (\ref{eq_N2S_constraint}).}
\begin{align}
  \matWnull^T \matC \matWnull
  &=
  \pmat{\matI_m & \matNull} \matP^T \matV^T
  \matC
  \matV \matP \pmat{\matI_m\\ \matNull}\\
  &=
  \pmat{\matI_m & \matNull} \matP^T
  \matLambda
  \matP \pmat{\matI_m\\ \matNull}\\ 
  &=
  \pmat{\matI_m & \matNull}
  \matLambda^*
  \pmat{\matI_m\\ \matNull}\\ 
  &=
  \hat{\matLambda}^*.
\end{align}
Therefore we introduce a term into the objective function where
off-diagonal elements in $\matWnull^T \matC \matWnull$ are pushed
towards zero.

\section{Modified Objective Function}\label{sec_objfct_mod}
%====================================

In the following, we suggest a modified objective function by
introducing an additional term, derive a learning rule, analyze its
fixed points, and study the stability of the fixed points indirectly
through the behavior of the objective function.

\subsection{Modified Objective Function}\label{sec_objfct_mod_1}
%---------------------------------------

We suggest the following modified objective function:
\begin{align}\label{eq_objfct_mod}
  J(\matW)
  =
  \quarter \left[
    (1 + \alpha) \summe{j=1}{m} \left(\vecw_j^T\matC\vecw_j\right)^2
    -
    \alpha \summe{j=1}{m}\summe{k=1}{m}\left(\vecw_j^T\matC\vecw_k\right)^2
  \right].
\end{align}
Again, we are interested in the local maxima of this function. The
first term of (\ref{eq_objfct_mod}) coincides with the original
objective function (\ref{eq_objfct_orig}). A second term with negative
sign is added which penalizes non-zero off-diagonal elements in
$\matW^T \matC \matW$ as motivated in section \ref{sec_objfct_orig}. A
weight factor $\alpha$ is introduced which expresses the influence of
the second term. The factors of the two terms are chosen such that
terms $(\vecw_j^T \matC \vecw_k)^2$ with $j = k$ are weighted with
$1$, and terms with $j \neq k$ are weighted with $-\alpha$. By writing
the equation in this way we can avoid the use of Kronecker's
delta. Due to the negative sign and the squared expressions, the terms
with $j \neq k$ are maximized if they are zero.

\subsection{Derivation of Modified Learning Rule}\label{sec_lr_mod}
%------------------------------------------------

To derive a learning rule from the modified objective function
(\ref{eq_objfct_mod}), we first determine its derivative with respect
to a single weight vector $\vecw_l$:
\begin{align}
  \nonumber
  &\ddf{J}{\vecw_l}\\
  &=
  \half \left[
    (1 + \alpha) \summe{j=1}{m} \left(\vecw_j^T\matC\vecw_j\right)
    \ddf{\vecw_j^T\matC\vecw_j}{\vecw_l}
    -
    \alpha \summe{j=1}{m}\summe{k=1}{m}\left(\vecw_j^T\matC\vecw_k\right)
    \ddf{\vecw_j^T\matC\vecw_k}{\vecw_l}
    \right]\\
  &=
  (1 + \alpha) \summe{j=1}{m} \left(\vecw_j^T\matC\vecw_j\right)
  \matC \vecw_j \delta_{jl}
  - \half \alpha \summe{j=1}{m}\summe{k=1}{m}\left(\vecw_j^T\matC\vecw_k\right)
  \left(\matC \vecw_k \delta_{jl} + \matC \vecw_j \delta_{kl}\right)\\
   &=
  (1 + \alpha) \left(\vecw_l^T\matC\vecw_l\right) \matC \vecw_l
  - \half \alpha \left[
    \summe{k=1}{m}\left(\vecw_l^T\matC\vecw_k\right) \matC \vecw_k
    +
    \summe{j=1}{m}\left(\vecw_j^T\matC\vecw_l\right) \matC \vecw_j
    \right]\\
   &=
  (1 + \alpha) \left(\vecw_l^T\matC\vecw_l\right) \matC \vecw_l
  - \half \alpha \left[
    \summe{j=1}{m}\left(\vecw_l^T\matC\vecw_j\right) \matC \vecw_j
    +
    \summe{j=1}{m}\left(\vecw_j^T\matC\vecw_l\right) \matC \vecw_j
    \right]\\
   &=
  (1 + \alpha) \left(\vecw_l^T\matC\vecw_l\right) \matC \vecw_l
  - \alpha \summe{j=1}{m}\left(\vecw_j^T\matC\vecw_l\right) \matC \vecw_j\\
   &=
  (1 + \alpha) \left(\vecw_l^T\matC\vecw_l\right) \matC \vecw_l
  - \alpha \summe{j=1}{m}\matC \vecw_j\vecw_j^T\matC\vecw_l\\
   &=
  (1 + \alpha) \left(\vecw_l^T\matC\vecw_l\right) \matC \vecw_l
  - \alpha \matC \left(\summe{j=1}{m} \vecw_j\vecw_j^T\right)\matC\vecw_l\\
   &=
  (1 + \alpha) \left(\vecw_l^T\matC\vecw_l\right) \matC \vecw_l
  - \alpha \matC \matW \matW^T \matC\vecw_l.
\end{align}
Now we combine the expression above into a derivative with respect to
the entire matrix $\matW$ (with $m$ columns $\vecw_l$, $l =
1,\ldots,m$):
\begin{align}
  \matM
  \coloneqq
  \ddf{J}{\matW}
  &=
  (1 + \alpha) \matC \matW
  \underbrace{\Diag{j=1}{m}\{\vecw_j^T \matC \vecw_j\}}_{\matD}
  - \alpha \matC \matW \matW^T \matC \matW\\
  &=
  (1 + \alpha) \matC \matW \matD
  - \alpha \matC \matW \matW^T \matC \matW.
\end{align}
In our previous work \cite[]{own_Moeller20a} we found that there are
two variants to eliminate the Lagrange multipliers, the first leading
to ``uninteresting'' principal subspace rules, the second to
``interesting'' PCA rules. We use the second variant and our ``short''
form derivation and obtain the following ``modified'' learning rule
which we henceforth refer to as ``M2S'':
\begin{align}
  \nonumber
  \tau \matWdot
  &=
  \matM\\
  &-
  \matW \matM^T \matW\\[5mm]
  \nonumber
  &=
  (1 + \alpha) \matC \matW \matD
  - \alpha \matC \matW \matW^T \matC \matW\\
  &-
  \matW
  \left[
    (1 + \alpha) \matD \matW^T \matC - \alpha \matW^T \matC \matW \matW^T \matC
  \right]
  \matW\\[5mm]
  \nonumber
  &=
  (1 + \alpha) \matC \matW \matD
  - \alpha \matC \matW (\matW^T \matC \matW)\\
  \label{eq_M2S_form0}
  &-
  (1 + \alpha) \matW \matD (\matW^T \matC \matW)
  + \alpha \matW (\matW^T \matC \matW) (\matW^T \matC \matW).
\end{align}
We can arrange equation (\ref{eq_M2S_form0}) in two ways. In the first
arrangement, we sort the terms according to the common factors
$(1+\alpha)$ and $-\alpha$:
\begin{align}\label{eq_M2S_form1}
  \tau \matWdot
  =
  (1 + \alpha) (\matC\matW\matD-\matW\matD\matW^T\matC\matW)
  - \alpha (\matC\matW-\matW\matW^T\matC\matW)(\matW^T\matC\matW).
\end{align}
This arrangement leads to an interesting insight on the fixed-point
structure of ``M2S'' which is elaborated in section \ref{sec_M2S_fp}.

The second arrangement is obtained from (\ref{eq_M2S_form0})
by combining the first with the second and the third with the forth
term, and factoring out common terms:
\begin{align}
  \tau \matWdot
  &=
  \matC \matW \left[(1 + \alpha) \matD - \alpha \matW^T\matC\matW\right]
  - \matW \left[(1 + \alpha) \matD - \alpha \matW^T\matC\matW\right]
  \matW^T\matC\matW\\
  \label{eq_M2S_form2}
  &=
   \matC \matW \matD'_\alpha
  - \matW \matD'_\alpha \matW^T\matC\matW.
\end{align}
We see that we obtain the same form as in ``N2S'' (\ref{eq_N2S}), but
with a matrix
\begin{align}\label{eq_Dp_alpha}
\matD'_\alpha = (1 + \alpha) \matD - \alpha \matW^T\matC\matW
\end{align}
instead of $\matD$. Note that $\matD'_\alpha$ is not generally
diagonal (but $\matDnull'_\alpha$ would be diagonal if the rule
actually converges to the principal eigenvectors).

\subsection{Fixed Points of Modified Learning Rule}\label{sec_M2S_fp}
%--------------------------------------------------

We can gain an interesting insight on the fixed-point structure of
``M2S'' from an analysis of the first arrangement of terms in
(\ref{eq_M2S_form1}). We see that the first term coincides with the
original learning rule ``N2S'' from (\ref{eq_N2S}). The second term
contains the right-hand side of Oja's subspace rule \exteq{110}
\begin{align}\label{eq_Oja_subspace}
\tau \matWdot = \matC\matW-\matW\matW^T\matC\matW
\end{align}
as the first factor \cite[]{nn_Oja89}. We know from \exteq{113} that
the fixed points of (\ref{eq_Oja_subspace}) are
\begin{align}\label{eq_Oja_subspace_fp}
\matWnull = \matV \matP \pmat{\matR\\ \matNull}
\end{align}
where $\matR$ is an arbitrary orthogonal matrix, thus the subspace
factor in the second term of (\ref{eq_M2S_form1}) will disappear as
soon as the eigenvector estimates span the same subspace as an
arbitrary selection of $m$ eigenvectors of $\matC$. The general
fixed-point solution of ``N2S'' (\ref{eq_N2S_fp_general}) always
fulfills (\ref{eq_Oja_subspace_fp}) with $\matR = \matU^{*T}\matP^*$
($\matU^*$ and $\matP^*$ are orthogonal, as is their product), thus
the second term of (\ref{eq_M2S_form1}) disappears in the fixed points
of ``N2S''. This leads to the insight that all fixed points of ``N2S''
are also present in ``M2S''. The additional term in the modified
objective function apparently only shapes the landscape outside the
fixed points. Note, however, that learning rule ``M2S'' may have
additional fixed points compared to ``N2S''.

Aside from this observation, the interpretation of the second term is
difficult. The entire second term may also disappear for other values
of $\matW$, depending on the interplay between first and second
factor. Moreover, the negative sign of the second term implies that
this term will probably not push $\matW$ towards the subspace
described above.

For the fixed-point analysis of ``M2S'', we proceed as in our previous
work \cite[]{own_Moeller20a}. We express $\matWnull$ through the
projections $\matAnull$ onto the eigenvectors by $\matWnull = \matV
\matAnull$, apply $\matV^T \matC \matV = \matLambda$, insert the
ansatz \exteq{84}
\begin{align}
\matAnull = \matQ \pmat{\matI_m\\ \matNull}
\end{align}
where $\matQ$ is an orthogonal matrix and therefore $\matAnull$ is
semi-orthogonal (located on a Stiefel manifold defined by $\matAnull^T
\matAnull = \matI_m$), and define
\begin{align}
  \matM
  \coloneqq
  \matQ^T \matLambda \matQ
  =
  \pmat{\matS & \matT^T\\ \matT & \matU}.
\end{align}
In appendix \ref{app_constraints_S_T} we describe two attempts ---
starting from either (\ref{eq_M2S_form1}) or (\ref{eq_M2S_form2}) ---
at deriving constraints on $\matS$ and $\matT$ which lead to the same
result, namely
\begin{align}
  \label{eq_M2S_constraint_S}
  \matS\matDnull
  &=
  \matDnull\matS\\
  \label{eq_M2S_constraint_T}
  \matT \left[(1+\alpha)\matDnull-\alpha\matS\right]
  &=
  \matNull.
\end{align}
While the constraint on $\matS$ (\ref{eq_M2S_constraint_S}) coincides
with the one for ``N2S'', the constraint on $\matT$
(\ref{eq_M2S_constraint_T}) differs from the one for ``N2S'' (where it
is $\matT\matDnull=\matNull$ with the only solution $\matT =
\matNull$). The constraint (\ref{eq_M2S_constraint_T}) also has the
solution $\matT = \matNull$, but can have additional, non-zero
solutions if the factor
$\matD'_\alpha=(1+\alpha)\matDnull-\alpha\matS$ is singular. A
simulation shows that $\det\{\matD'_{\alpha}\}$ can actually be zero,
see figure \ref{fig_det_Dp_alpha} in appendix \ref{app_add_fp}. We
will focus on the case $\matT = \matNull$ which coincides with
``N2S''. For this case, the derivation completely coincides with the
one of ``N2S'' (from equation \exteq{248} onward) and leads to the
special solution (\ref{eq_N2S_fp_special}) and the general solution
(\ref{eq_N2S_fp_general}).

\subsection{Stability Analysis}\label{sec_stability_mod}
%------------------------------

The stability analysis uses the same indirect approach as in our
previous work \cite[][Sec. 8]{own_Moeller20a}. We can use the the
following expressions from \exteq{330}, \exteq{335}, and \exteq{338}:
\begin{align}
  \matWnull^T \matC \matWnull
  &= \matU_m^T \hat{\matLambda}^* \matU_m \eqqcolon \matH\\
  \matW^T \matC \matW
  &=
  \matF^T \matH \matF + \matB^T \check{\matLambda}^* \matB\\
  \nonumber
  \matF^T \matH \matF
  &\approx
    \matH
  + \matA^T \matH
  + \matH \matA
  + \matA^T \matH \matA\\
  &- \half (\matA^T \matA + \matB^T \matB) \matH
  - \half \matH (\matA^T \matA + \matB^T \matB).
\end{align}
We compute the change in the objective function under a small step
from fixed point $\matWnull$ (on the Stiefel manifold) to point
$\matW$ obtained by an approximated back-projection onto the Stiefel
manifold. The step is parametrized by a skew-symmetric $m \times m$
matrix $\matA$ and an $(n-m)\times m$ matrix $\matB$. For the modified
objective function from equation (\ref{eq_objfct_mod}) we get
\begin{align}
  \Delta J
  &=
  J(\matW) - J(\matWnull)\\
  &=
  (1+\alpha) \underbrace{\quarter \bigg[
    \summe{j=1}{m} \left(\vecw_j^T\matC\vecw_j\right)^2
    -
    \summe{j=1}{m} \left(\vecwnull_j^T\matC\vecwnull_j\right)^2
    \bigg]}_{\Delta J_1}\\
  &+ \alpha \underbrace{\quarter \bigg[
    \summe{j=1}{m}\summe{k=1}{m}\left(\vecwnull_j^T\matC\vecwnull_k\right)^2
    -
    \summe{j=1}{m}\summe{k=1}{m}\left(\vecw_j^T\matC\vecw_k\right)^2
    \bigg]}_{\Delta J_2}
\end{align}
where the negative sign was incorporated into $\Delta J_2$. We see
that $\Delta J_1$ describes the change of the original objective
function (\ref{eq_objfct_orig}) for which we derived \exteq{413}
\begin{align}
  \nonumber
  &\Delta J_1\approx\\
  &
  \half \summe{j=1}{m} \matH_{jj} \left\{
  (\matA^T \matH \matA)_{jj}
  - [(\matA^T \matA + \matB^T \matB) \matH]_{jj}
  + (\matB^T \check{\matLambda}^* \matB)_{jj}
  \right\}
  + \summe{j=1}{m} [(\matA^T \matH)_{jj}]^2.
\end{align}
For $\Delta J_2$ we obtain
\begin{align}
  \nonumber
  \Delta J_2
  &=
  \quarter \summe{j=1}{m} \summe{k=1}{m}
  \left[
    (\vecwnull_j^T\matC\vecwnull_k)^2 -
    (\vecw_j^T\matC\vecw_k)^2
    \right]\\
  &= 
  \quarter \summe{j=1}{m} \summe{k=1}{m}
  \left[
    (\vece_j^T\matWnull^T\matC\matWnull\vece_k)^2 -
    (\vece_j^T\matW^T\matC\matW\vece_k)^2
    \right]\\
  &=
  \quarter \summe{j=1}{m} \summe{k=1}{m}
  \left[
    (\vece_j^T\matH\vece_k)^2 -
    (\vece_j^T\{\matF^T\matH\matF + \matB^T\check{\matLambda}^*\matB\}\vece_k)^2
    \right]\\
  &=
  \quarter \left[
  \|\matH\|^2_F
  - \|\matF^T\matH\matF + \matB^T\check{\matLambda}^*\matB\|^2_F\right]\\
  &=
  \quarter \left[
  \|\matU_m^T \hat{\matLambda}^* \matU_m \|^2_F
  - \tr\{(\matF^T\matH\matF + \matB^T\check{\matLambda}^*\matB)^2\}\right]\\
  &=
  \quarter \left[
  \|\hat{\matLambda}^*\|^2_F
  - \tr\{(\matF^T\matH\matF + \matB^T\check{\matLambda}^*\matB)^2\}\right]\\
  &\approx
  \quarter \left[
  \|\hat{\matLambda}^*\|^2_F
  - \tr\{
  (\matF^T\matH\matF)^2
  + 2 (\matF^T\matH\matF)(\matB^T\check{\matLambda}^*\matB)
  \}\right]\\
  \label{eq_deltaJ2}
  &=
  \quarter \left[
  \|\hat{\matLambda}^*\|^2_F
  - \tr\{(\matF^T\matH\matF)^2\}
  - 2 \tr\{(\matF^T\matH\matF)(\matB^T\check{\matLambda}^*\matB)\}\right]
\end{align}
where we omitted terms above second order in the approximation. Note
that for symmetric $\matX$ we have $\|\matX\|^2_F =
\tr\{\matX^T\matX\} = \tr\{\matX^2\}$.

We further process the second term of (\ref{eq_deltaJ2}), using the
invariance of the trace to cyclic permutation, exploiting skew-symmetry
$\matA^T = -\matA$ and symmetry $\matH^T = \matH$, and omitting terms
above second order in $\matA$ and $\matB$:
\begin{align}
  \nonumber
  &
  \tr\{(\matF^T\matH\matF)^2\}\\[5mm]
  \nonumber
  &\approx
  \tr\bigg\{
  \Big[\matH
  + \matA^T \matH
  + \matH \matA
  + \matA^T \matH \matA\\
  &- \half (\matA^T \matA + \matB^T \matB) \matH
  - \half \matH (\matA^T \matA + \matB^T \matB)
  \Big]^2\bigg\}\\[5mm]
  \nonumber
  &\approx
  \tr\bigg\{
  \matH^2 + \matH\matA^T\matH + \matH^2\matA + \matH\matA^T\matH\matA\\
  \nonumber
  &-\half\matH(\matA^T\matA+\matB^T\matB)\matH
  -\half\matH^2(\matA^T\matA+\matB^T\matB)\\
  \nonumber
  &+\matA^T\matH^2 + \matA^T\matH\matA^T\matH + \matA^T\matH\matH\matA\\
  \nonumber
  &+\matH\matA\matH + \matH\matA\matA^T\matH + \matH\matA\matH\matA\\
  \nonumber
  &+\matA^T\matH\matA\matH\\
  &-\half(\matA^T\matA+\matB^T\matB)\matH^2
  -\half \matH(\matA^T\matA+\matB^T\matB)\matH
  \bigg\}\\[5mm]
  \nonumber
  &=
  \tr\bigg\{
  \matH^2 + \matH^2\matA^T + \matH^2\matA + \matH\matA^T\matH\matA\\
  \nonumber
  &-\half\matH^2(\matA^T\matA+\matB^T\matB)
  -\half\matH^2(\matA^T\matA+\matB^T\matB)\\
  \nonumber
  &+\matH^2\matA^T + \matH\matA^T\matH\matA^T + \matH^2\matA\matA^T\\
  \nonumber
  &+\matH^2\matA + \matH^2\matA\matA^T + \matH\matA\matH\matA\\
  \nonumber
  &+\matH\matA^T\matH\matA\\
  &-\half\matH^2(\matA^T\matA+\matB^T\matB)
  -\half\matH^2(\matA^T\matA+\matB^T\matB)\bigg\}\\[5mm]
  &=
  \tr\left\{
  \matH^2
  -2\matH^2(\matA^T\matA+\matB^T\matB)
  +2\matH^2\matA\matA^T\right\}\\[5mm]
  &=
  \tr\left\{
  \matH^2
  -2\matH^2(\matA^T\matA+\matB^T\matB)
  +2\matH^2\matA^T\matA\right\}\\[5mm]
  &=
  \tr\left\{
  \matH^2
  -2\matH^2\matB^T\matB\right\}\\[5mm]
  &=
  \tr\left\{\matH^2\right\}
  -2\tr\left\{\matH^T\matB^T\matB\matH\right\}\\[5mm]
  &=
  \|\hat{\matLambda}^*\|^2_F
  -2\|\matB\matH\|^2_F.
\end{align}
The third term of (\ref{eq_deltaJ2}) only has terms of second order
(or below) by taking $\matH$ from the first factor:
\begin{align}
  \tr\left\{(\matF^T\matH\matF)(\matB^T\check{\matLambda}^*\matB)\right\}
  \approx
  \tr\left\{\matH\matB^T\check{\matLambda}^*\matB\right\}.
\end{align}
We summarize:
\begin{align}
  \label{eq_deltaJ2_final}
  \Delta J_2
  \approx
  \half\left[
    \tr\{\matH^2\matB^T\matB\} - \tr\{\matH\matB^T\check{\matLambda}^*\matB\}
    \right].
\end{align}
For the special case with pairwise different elements in $\matDnull$
we have $\matU_m = \matI_m$ and thus $\matH = \hat{\matLambda}^*$. We
apply \exteq{576} and obtain
\begin{align}
  \Delta J_2
  &\approx
  \half\left[
    \tr\{\hat{\matLambda}^{*2}\matB^T\matB\}
    -
    \tr\{\hat{\matLambda}^*\matB^T\check{\matLambda}^*\matB\}
    \right]\\
  &=
  \half\left[
    \summe{j=1}{m} \hat{\lambda}_j^{*2} (\matB^T\matB)_{jj}
    -
    \summe{j=1}{m} \hat{\lambda}_j^* (\matB^T\check{\matLambda}^*\matB)_{jj}
    \right]\\
  &=
  -\half\left[
    \summe{j=1}{m} \hat{\lambda}_j^* (\matB^T\check{\matLambda}^*\matB)_{jj}
    -
    \summe{j=1}{m} \hat{\lambda}_j^{*2} (\matB^T\matB)_{jj}
    \right]\\
  \label{eq_deltaJ2_final_mod}
  &=
   -\half\summe{j=1}{m}\left[
     \hat{\lambda}_j^* (\matB^T\check{\matLambda}^*\matB)_{jj}
    -
    \hat{\lambda}_j^{*2} (\matB^T\matB)_{jj}
    \right].
\end{align}
For the special case we also have with \exteq{376}
\begin{align}
  \nonumber
  \Delta J_1
  &\approx
  \half\summe{j=1}{m}\left[
    \hat{\lambda}^*_j  (\matA^T \hat{\matLambda}^* \matA)_{jj}
    - \hat{\lambda}^{*^2}_j (\matA^T \matA)_{jj}\right]\\
  &+
  \half\summe{j=1}{m}\left[
    \hat{\lambda}^*_j (\matB^T \check{\matLambda}^* \matB)_{jj}
    - \hat{\lambda}^{*^2}_j (\matB^T \matB)_{jj}\right],
\end{align}
thus by combining the two expressions we obtain
\begin{align}
  \nonumber
  \Delta J
  &\approx
    \half (1 + \alpha) \summe{j=1}{m}\left[
    \hat{\lambda}^*_j  (\matA^T \hat{\matLambda}^* \matA)_{jj}
    - \hat{\lambda}^{*^2}_j (\matA^T \matA)_{jj}\right]\\
  \nonumber
  &+
  \half (1 + \alpha) \summe{j=1}{m}\left[
    \hat{\lambda}^*_j (\matB^T \check{\matLambda}^* \matB)_{jj}
    - \hat{\lambda}^{*^2}_j (\matB^T \matB)_{jj}\right]\\
  &-
  \half \alpha \summe{j=1}{m}\left[
     \hat{\lambda}_j^* (\matB^T\check{\matLambda}^*\matB)_{jj}
    -
    \hat{\lambda}_j^{*2} (\matB^T\matB)_{jj}
    \right]\\
  \nonumber
  &=
     \half (1 + \alpha) \summe{j=1}{m}\left[
    \hat{\lambda}^*_j  (\matA^T \hat{\matLambda}^* \matA)_{jj}
    - \hat{\lambda}^{*^2}_j (\matA^T \matA)_{jj}\right]\\
  &+
  \half \summe{j=1}{m}\left[
    \hat{\lambda}^*_j (\matB^T \check{\matLambda}^* \matB)_{jj}
    - \hat{\lambda}^{*^2}_j (\matB^T \matB)_{jj}\right].
\end{align}
As in the original objective function (\ref{eq_objfct_orig}), we can
demonstrate the existence of a maximum ($\Delta J < 0$) if the first
$m$ eigenvectors are associated with the $m$ largest eigenvalues
\exteq{section 8.4.1}; these are the ``desired'' fixed
points. Otherwise we obtain a saddle point or a minimum ($\Delta J >
0$ in some directions).

For the general case where diagonal elements in $\matDnull$ may
coincide and where we have $\matH = \matU_m^T \hat{\matLambda}^*
\matU_m$, we could show that $\Delta J > 0$ for $\matB = \matNull$ and
a specific choice of $\matA$ \exteq{section 8.4.2}. Since $\Delta J_2$
only depends on $\matB$ and disappears for $\matB = \matNull$, we can
demonstrate that the ``undesired'' fixed points are either saddle
points or minima.

We conclude that the additional term introduced in the modified
objective function (\ref{eq_objfct_mod}) leaves the stability of the
fixed points unchanged. We also see that the factor $(1+\alpha)$ leads
to a steeper shape of the objective function in the vicinity of the
fixed points, at least in some directions (determined by step
parameter $\matA$).

%#############################################################################
%#############################################################################
%#############################################################################

\section{Simulations}\label{sec_simulations}
%====================

As in our previous work, we restrict our simulations to averaged
learning rules operating on the covariance matrix $\matC =
E\{\vecx\vecx^T\}$ (in contrast, online learning rules operate on
individual data vectors $\vecx$).

\subsection{Methods}\label{sec_simulations_methods}
%-------------------

We explore the behavior of the following learning rules:
\begin{description}
\item[``TwJ2S''] from \exteq{449}, which is the same as rule (15a)
  from \cite{nn_Xu93}, with $\matTheta = \diag_{j=1}^m\{j/m\}$,
\item[``N2S''] from (\ref{eq_N2S}), which is the same as ``M2S'' with
  $\alpha=0$, and
\item[``M2S''] from (\ref{eq_M2S_form2}) for $\alpha \in \{1.0, 2.0,
  5.0, 10.0, 20.0\}$.
\end{description}
We determine eigenvector estimates $\matW$ with $n = 10$ and $m =
4$. We start from a random initial $\matW$ located on the Stiefel
manifold ($\matW^T \matW = \matI_m$) which is the same for all
learning rules and all figures.

We generate a $n \times n$ covariance matrix $\matC$ from a random
orthogonal $\matV$ and a diagonal eigenvalue matrix $\matLambda$
through $\matC = \matV \matLambda \matV^T$. The matrix $\matLambda$ is
generated from one of the following eigenvalue sets, either
\begin{description}
  \item[``nearby eigenvalues''] $\{0.91, 0.9, 0.8, \ldots, 0.1\}$ or
  \item[``evenly spaced eigenvalues''] $\{1.0, 0.9, 0.8, \ldots, 0.1\}$
\end{description}
through $\matLambda = \diag_{i=1}^n \lambda_i$.

The simulation uses an Euler step $\matW'_{t+1} = \matW_t +
\matWdot_t$ where $\matWdot_t$ contains the parameter $\gamma =
1/\tau$. Three different subsequent back-projection modes are tested:
\begin{description}
\item[``exact back-projection to Stiefel manifold'']
  \begin{align}
    \matW_{t+1} = \matW'_{t+1} \left( \matW'^T_{t+1}  \matW'_{t+1} \right)^{-\half},
  \end{align}
\item[``approximated back-projection to Stiefel manifold''] from \exteq{630}
  \begin{align}
    \matW_{t+1} = \matW'_{t+1} - \half \matW_t \matWdot_t^T \matWdot_t,
  \end{align}
\item[``no back-projection'']
  \begin{align}
    \matW_{t+1} = \matW'_{t+1}.
  \end{align}
\end{description}

To evaluate the deviation of $\matW$ from semi-orthogonality
(``orthonormality'' for short) and the deviation of $\matW$ from the
true principal eigenvectors (in arbitrary order), we define three
error measures $e_1$, $e_2$, and $e'_2$ on square matrices of size
$m$:
\begin{align}
  e_1(\matX)
  &=
  \frac{1}{m^2} \summe{i=1}{m}\summe{j=1}{m} \left|X_{ij} - \delta_{ij}\right|\\
  %-----
  e_2(\matX)
  &=
  \frac{1}{m} \summe{j=1}{m} \left|\max_{i=1}^m\{|(\vecx_j)_i|\} - 1\right|\\
  %-----
  e'_2(\matX)
  &=
  \half\left(e_2(\matX) + e_2(\matX^T)\right)
\end{align}
Error measure $e_1$ is zero if $\matX$ coincides with the identity
matrix of the same size. Error measure $e_2$ is zero if the maximal
absolute element in each column of $\matX$ is $1$. Error measure
$e'_2$ considers $e_2$ in both columns and rows. We define the
orthonormality error $e_o$ and the error of the projection to the
eigenvectors $e_p$ as
\begin{align}
  e_o(\matW) &= e_1(\matW^T\matW)\\
  e_p(\matW,\hat{\matV}) &= e'_2(\hat{\matV}^T\matW)
\end{align}
where $\hat{\matV}$ (size $n \times m$) contains the $m$ principal
eigenvectors in its columns. Error measure $e_o$ is zero for a
semi-orthogonal $\matW$. Error measure $e_p$ is zero if each
eigenvector estimate $\vecw_j$ corresponds to a true eigenvector
$\pm\vecv_i$ (arbitrary sign) in a one-to-one mapping. To motivate the
error measure $e_p$, we show two examples of final values of
$\hat{\matV}^T \matW$ which lead to $e_p \approx 0$. The first is from
learning rule ``TwJ2S'' where the ordering of the estimated
eigenvectors with respect to the eigenvalues is determined by the
fixed matrix $\matTheta$:
\begin{align}
  \hat{\matV}^T\matW =
  \begin{pmatrix*}[r]
       0.00 &  0.00 &  0.00 & -1.00\\
      -0.00 & -0.00 &  1.00 &  0.00\\
       0.00 & -1.00 & -0.00 & -0.00\\
       1.00 &  0.00 &  0.00 &  0.00
  \end{pmatrix*}.
\end{align}
The corresponding eigenvalue estimates $\vecw_j^T\matC\vecw_j$ are, in
the same order: $0.70,\, 0.80,\, 0.90,\, 1.00$. The second example is
from learning rule ``N2S'' where the approached ordering is arbitrary:
\begin{align}
  \hat{\matV}^T\matW =
  \begin{pmatrix*}[r]
    -0.00 & -1.00 &  0.00 &  0.00\\
     0.00 &  0.00 & -0.00 &  1.00\\
     1.00 & -0.00 &  0.00 & -0.00\\
    -0.00 &  0.00 &  1.00 &  0.00\\
  \end{pmatrix*}.
\end{align}
The eigenvalue estimates are, in the same order: $0.80,\, 1.00,\,
0.70,\, 0.90$.

\subsection{Results}\label{sec_simulations_results}
%-------------------

\begin{figure}[tp]
  \begin{center}
    
    \begin{subfigure}[t]{\textwidth}
      \begin{center}
        \includegraphics[width=74mm]{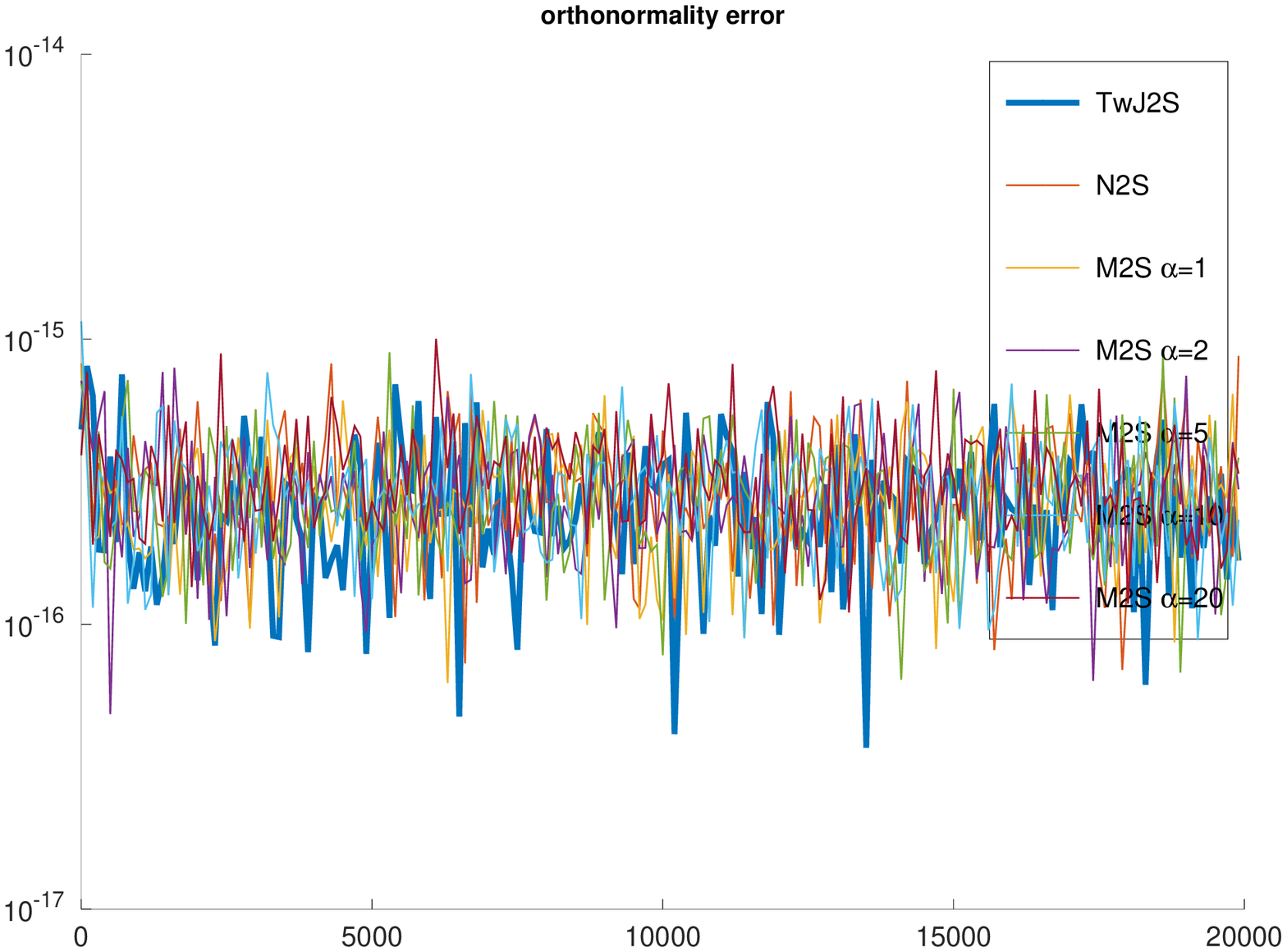}
        \includegraphics[width=74mm]{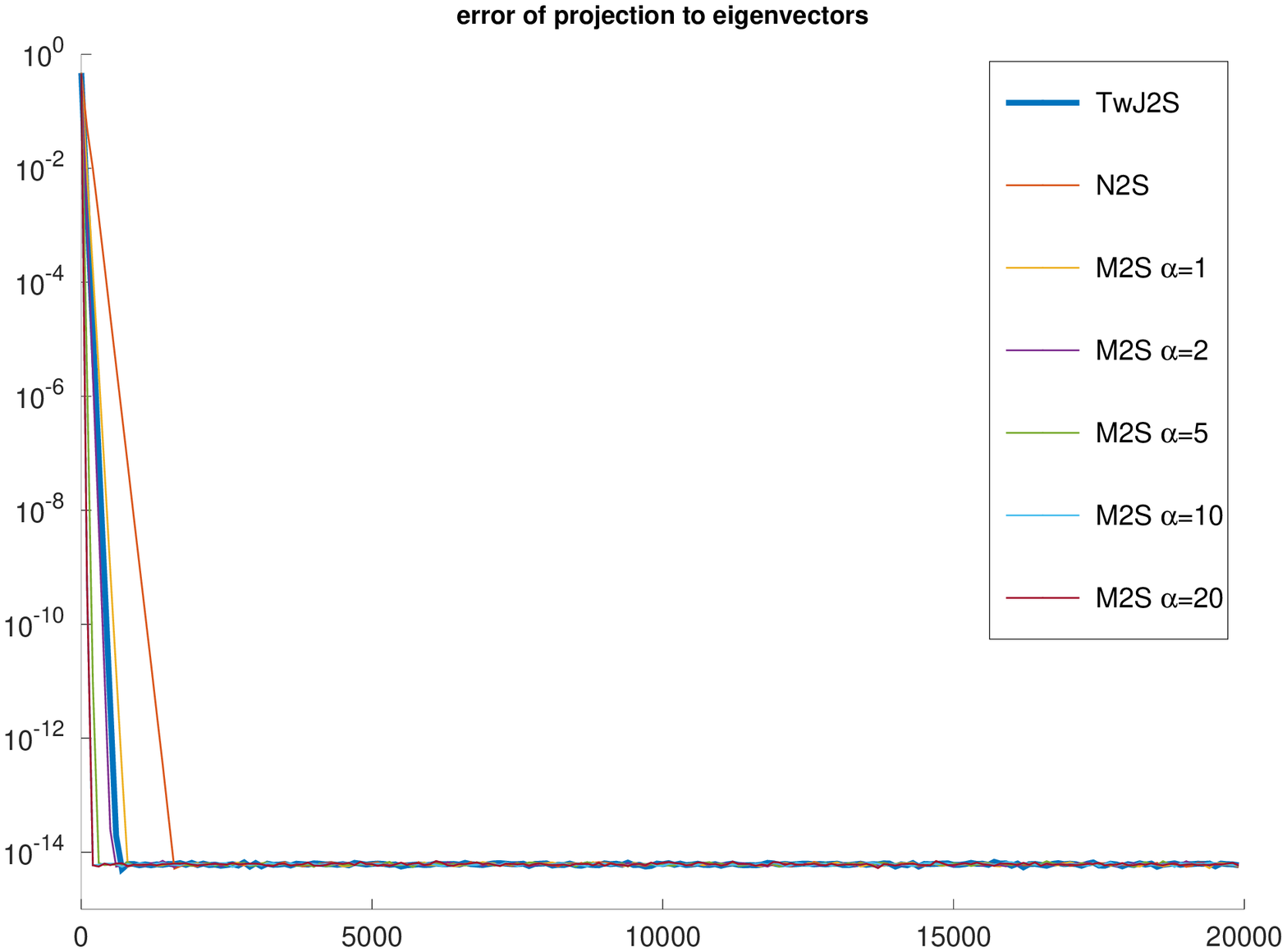}\\[5mm]
        \caption{Exact back-projection, $\gamma = 1$.}
        \label{fig_spaced_stepSt}
      \end{center}
    \end{subfigure}

    \begin{subfigure}[t]{\textwidth}
      \begin{center}
        \includegraphics[width=74mm]{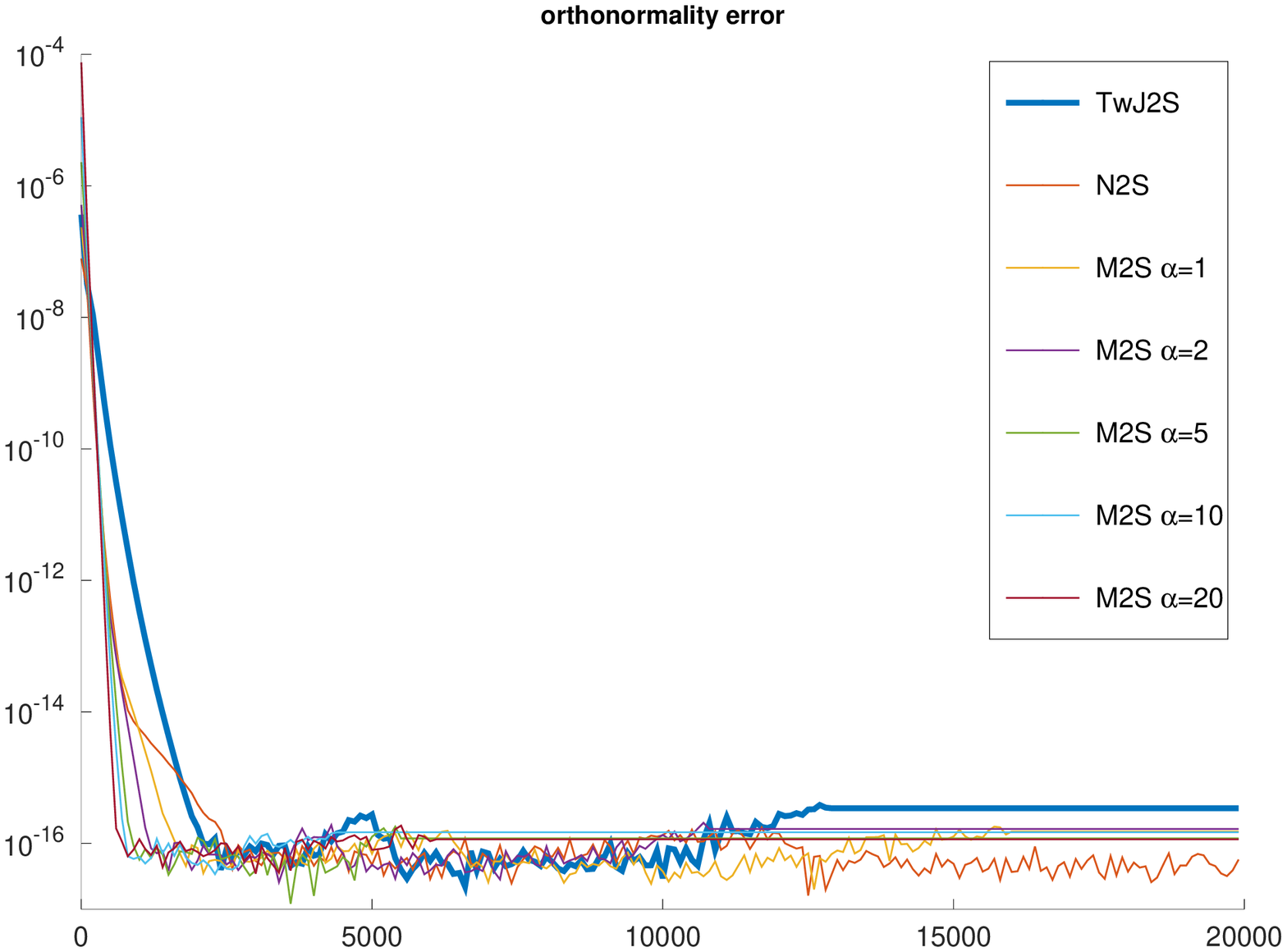}
        \includegraphics[width=74mm]{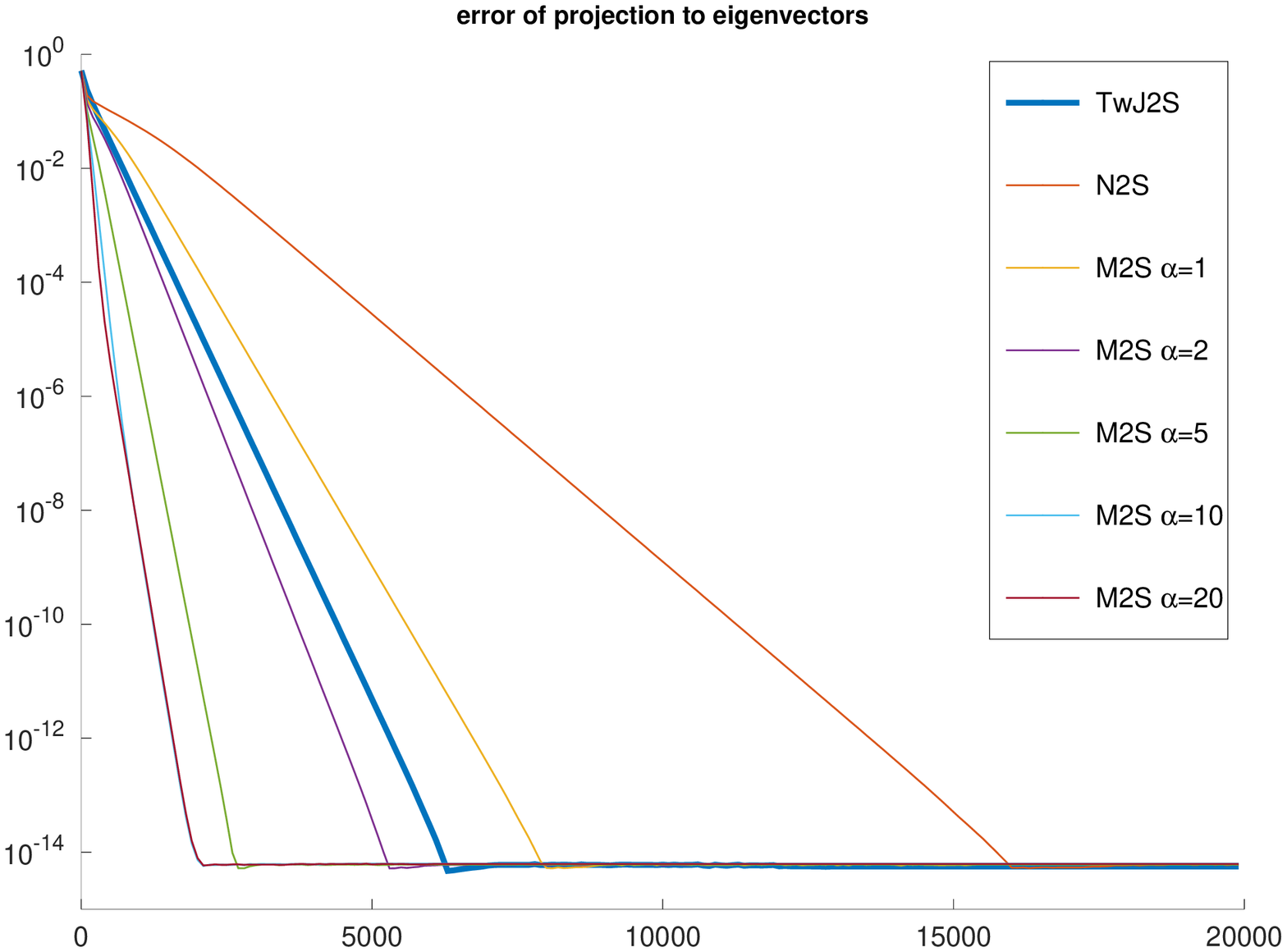}\\[5mm]
        \caption{Approximated back-projection, $\gamma = 0.1$.}
        \label{fig_spaced_stepStAppr}
      \end{center}
    \end{subfigure}

    \begin{subfigure}[t]{\textwidth}
      \begin{center}
        \includegraphics[width=74mm]{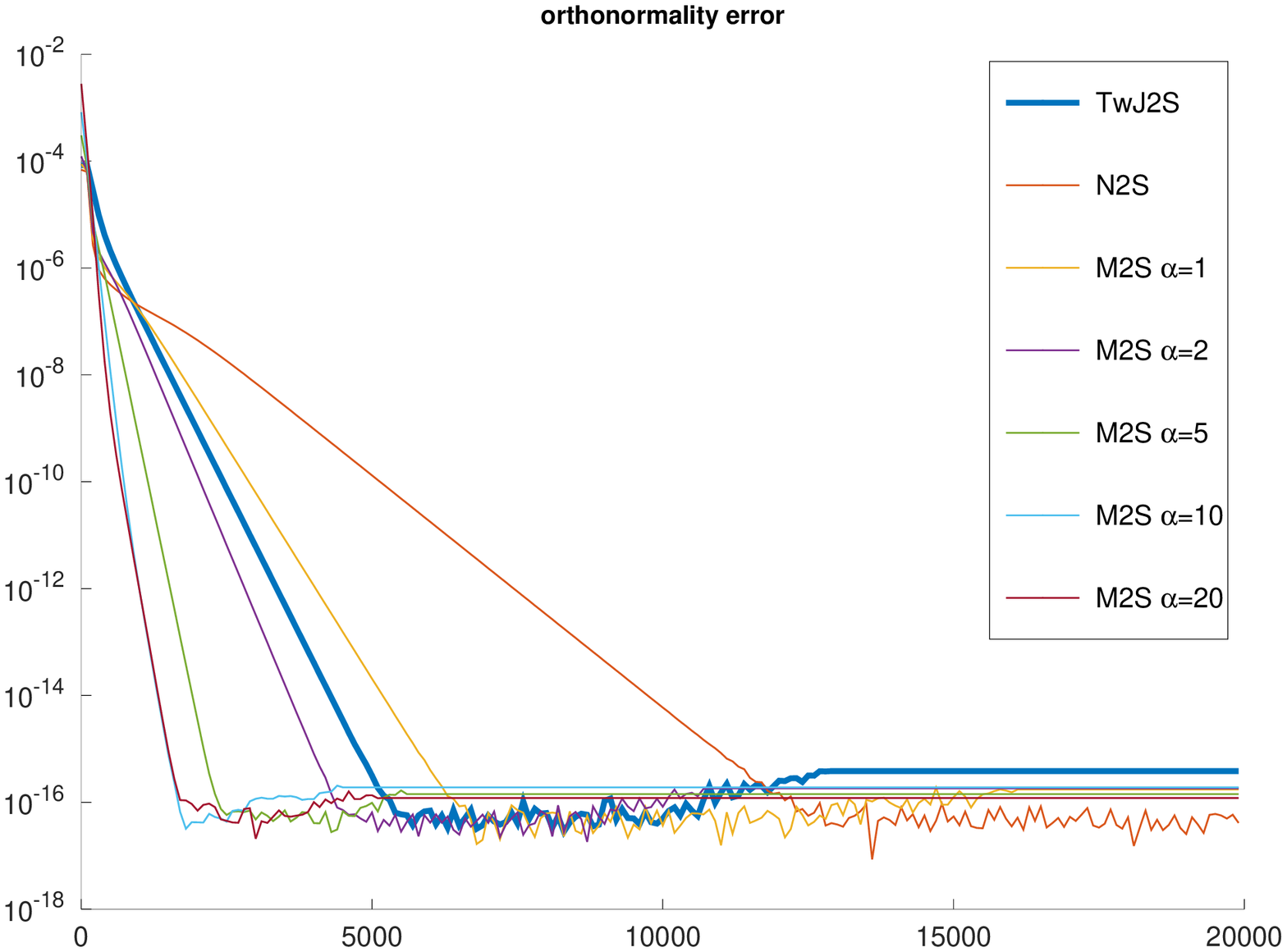}
        \includegraphics[width=74mm]{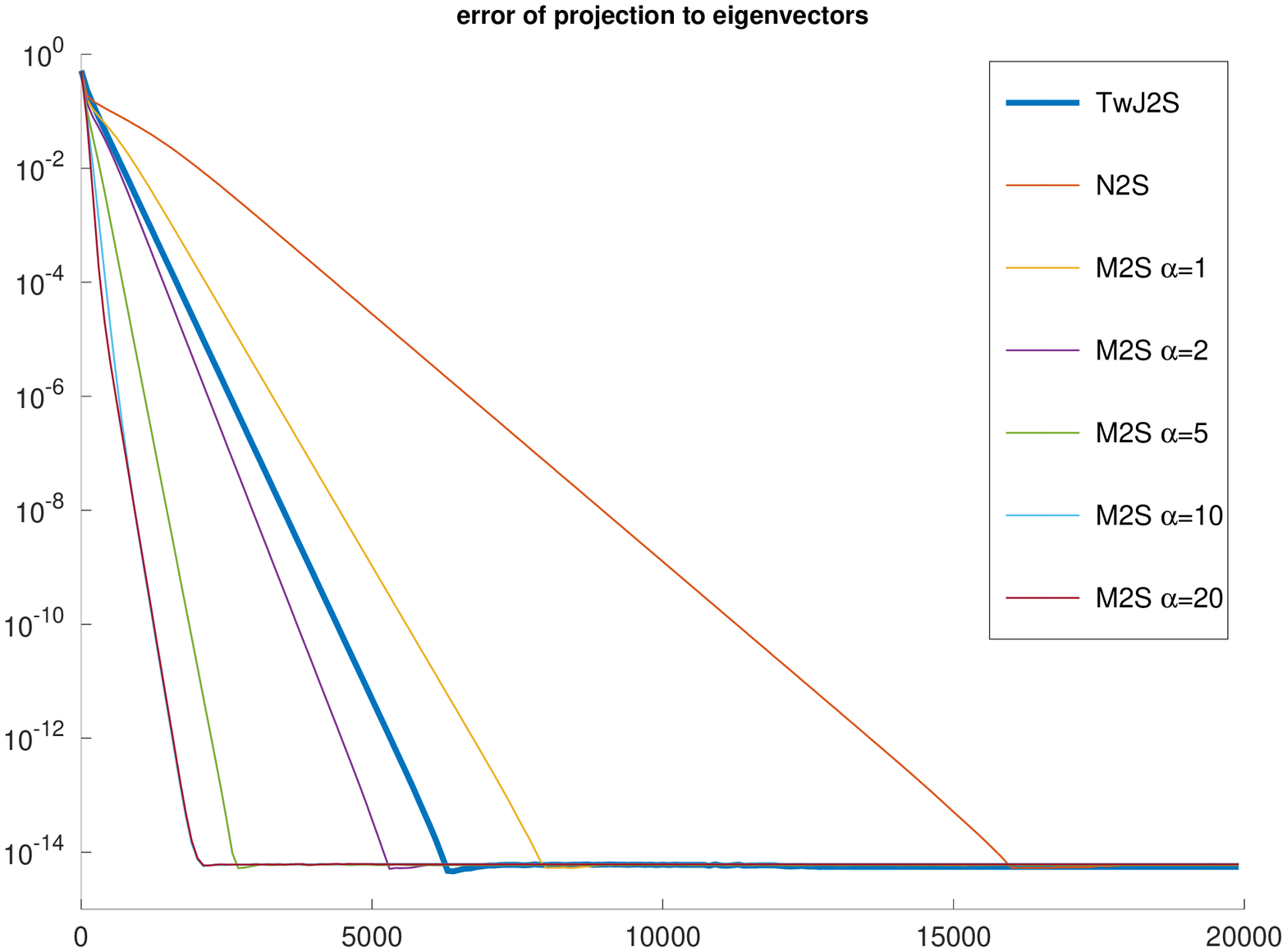}\\[5mm]
        \caption{No back-projection, $\gamma = 0.1$.}
        \label{fig_spaced_stepNone}
      \end{center}
    \end{subfigure}
 
    \caption{Orthonormality error $e_o$ (left) and error of projection
      to eigenvectors $e_p$ (right) for evenly spaced eigenvalues
      (logarithmic, $20.000$ steps, subsampling $100$).}
    \label{fig_spaced}
  \end{center}
\end{figure}

Figure \ref{fig_spaced} shows the simulation results for the evenly
spaced eigenvector set for the three back-projection methods (note the
reduced number of simulation steps). Looking at the projection error
$e_p$ (right diagrams), we see fast convergence for all learning
rules, particularly for the exact back-projection where the learning
rate $\gamma$ can be higher than in the other two back-projection
methods. ``N2S'' converges more slowly than ``TwJ2S'', but is in the
same convergence range. ``M2S'' converges faster with increasing
$\alpha$ and even surpasses ``TwJ2S'' (but see section
\ref{sec_discussion}), but the gain decreases for the highest values
of $\alpha$. The orthonormality error $e_o$ (left diagrams) stays
small for exact back-projection, reduces very fast for approximated
back-projection, and reduces somewhat slower for no
back-projection. In the latter two cases, increasing $\alpha$
accelerates the convergence of the orthonormality error. The improved
convergence of the orthonormality error from no back-projection to
approximated back-projection is not reflected in faster reduction of
the projection error, though.

\begin{figure}[tp]
  \begin{center}

    \begin{subfigure}[t]{\textwidth}
      \begin{center}    
        \includegraphics[width=74mm]{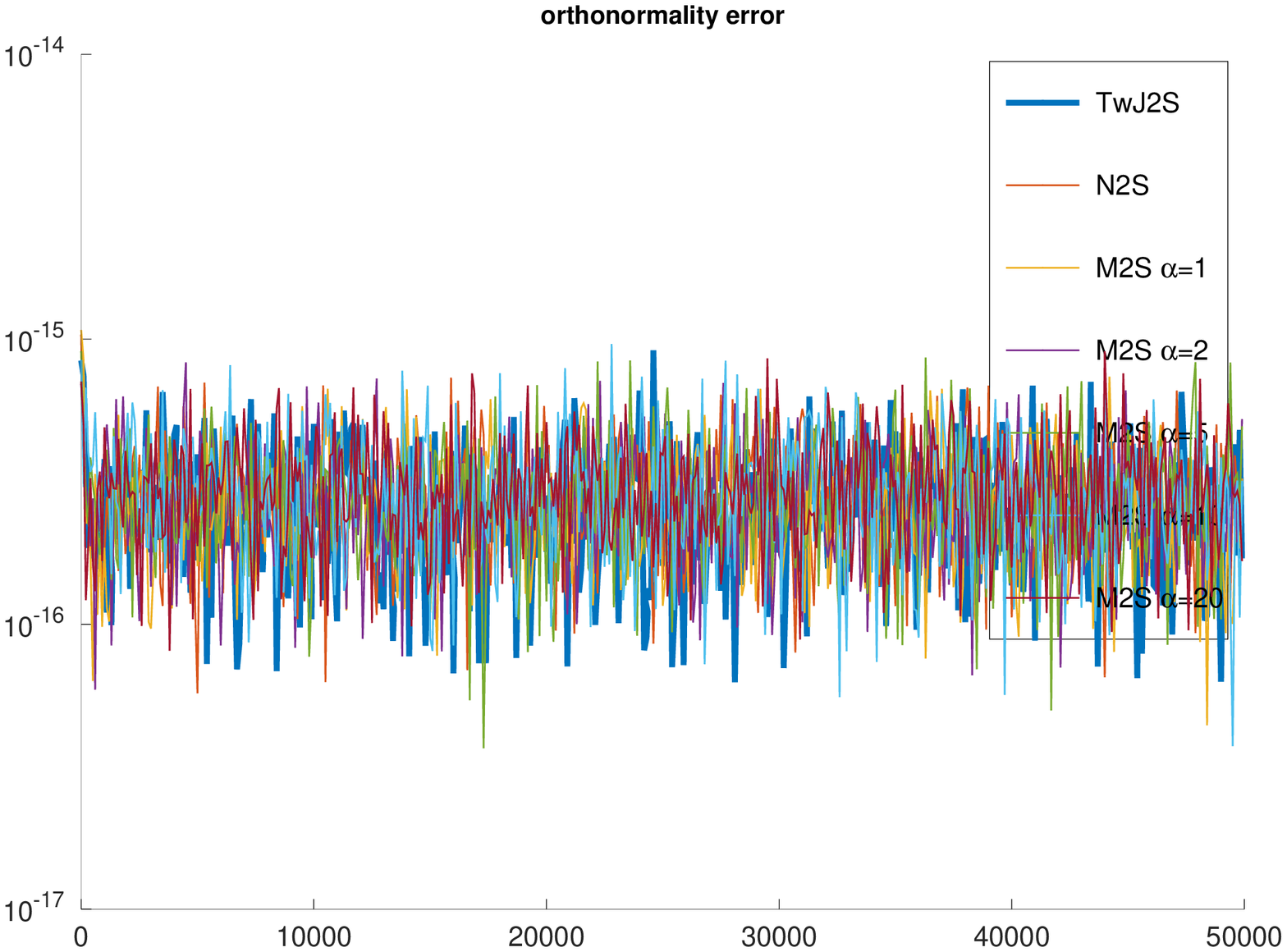}
        \includegraphics[width=74mm]{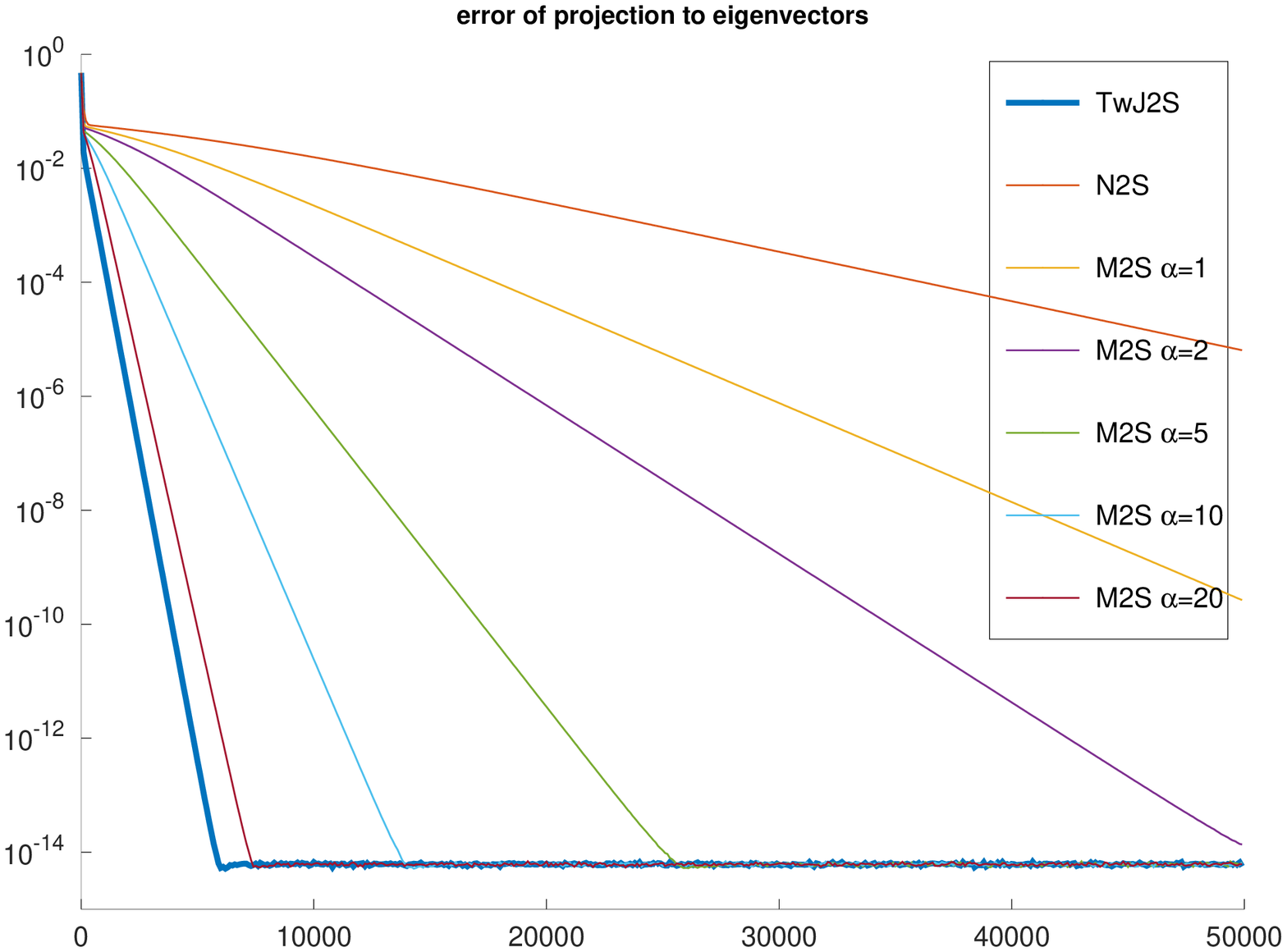}\\[5mm]
        \caption{Exact back-projection, $\gamma = 1$.}
        \label{fig_nearby_stepSt}
      \end{center}
    \end{subfigure}

    \begin{subfigure}[t]{\textwidth}
      \begin{center}
        \includegraphics[width=74mm]{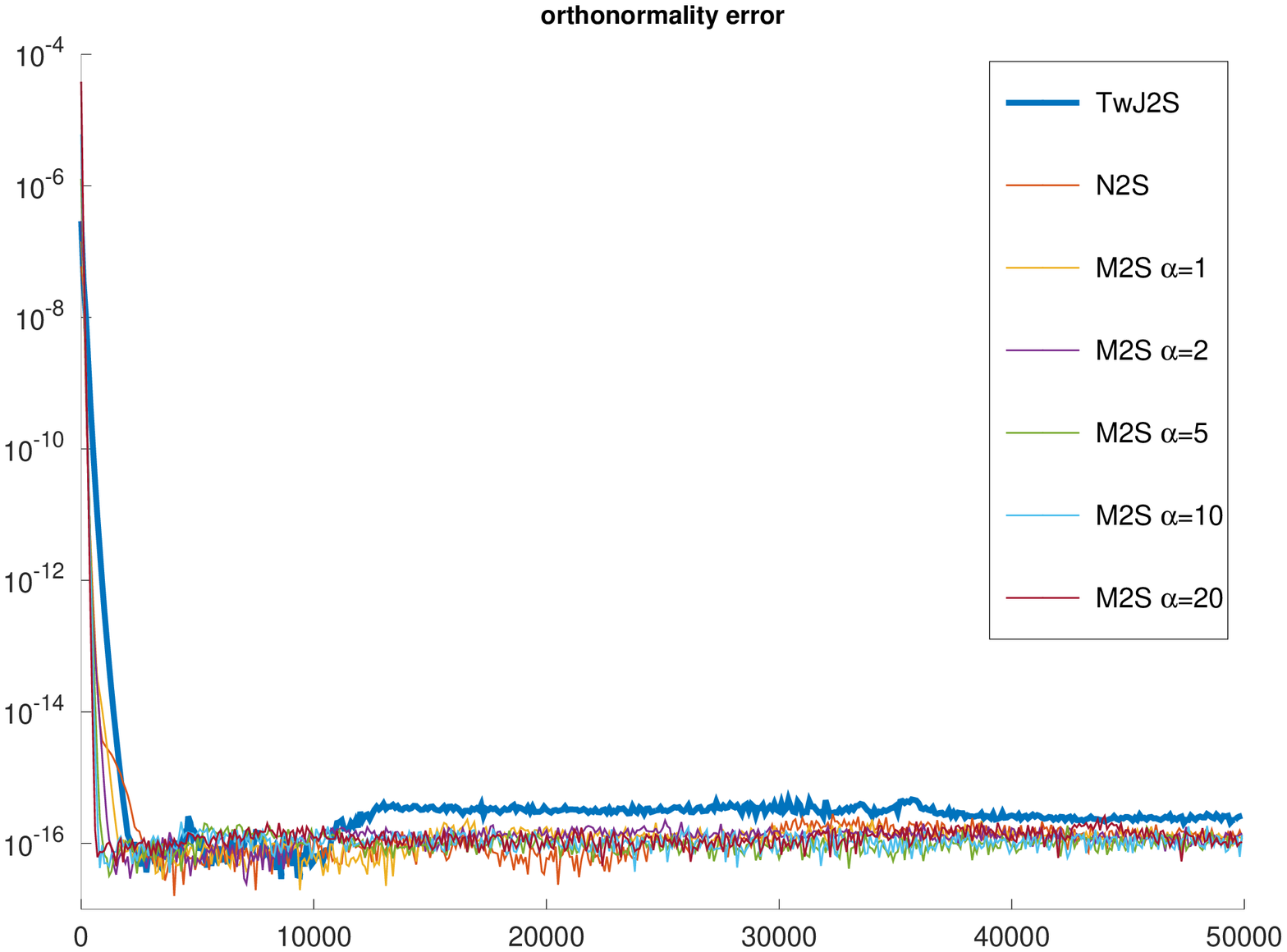}
        \includegraphics[width=74mm]{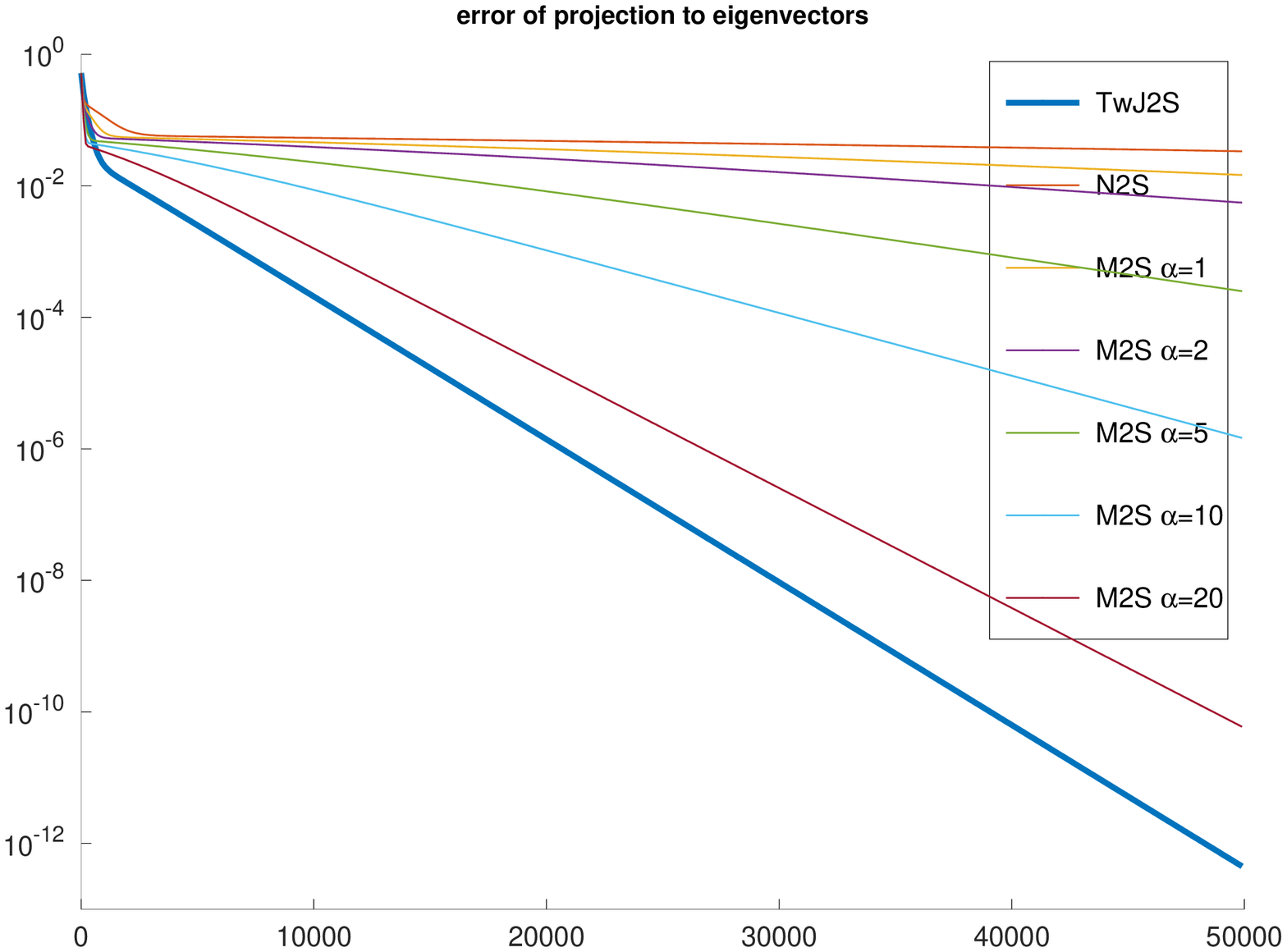}\\[5mm]
        \caption{Approximated back-projection, $\gamma = 0.1$.}
        \label{fig_nearby_stepStAppr}
      \end{center}
    \end{subfigure}

    \begin{subfigure}[t]{\textwidth}
      \begin{center}
        \includegraphics[width=74mm]{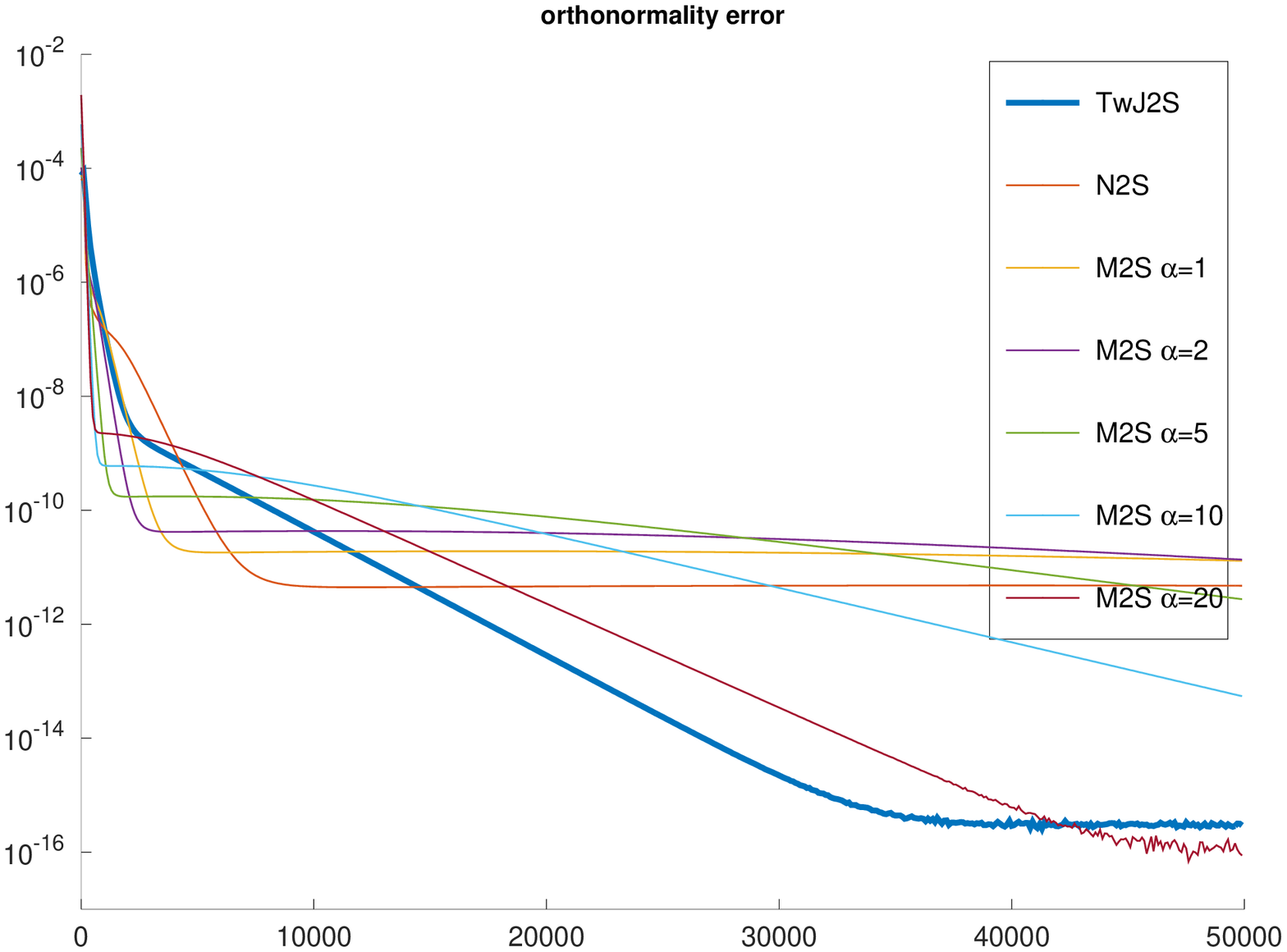}
        \includegraphics[width=74mm]{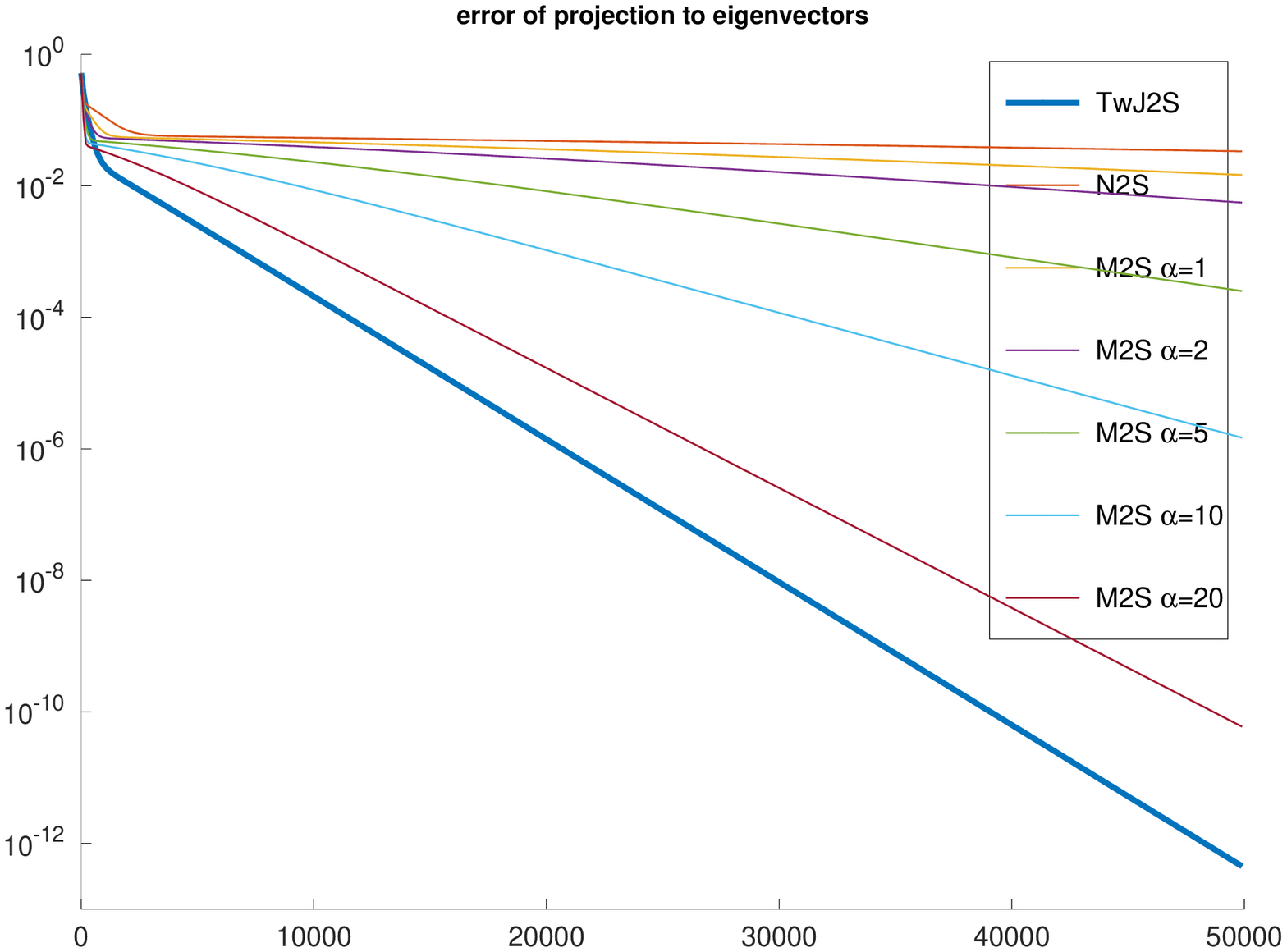}\\[5mm]
        \caption{No back-projection, $\gamma = 0.1$}
        \label{fig_nearby_stepNone}
      \end{center}
    \end{subfigure}

    \caption{Orthonormality error $e_o$ (left) and error of projection
      to eigenvectors $e_p$ (right) for nearby eigenvalues $\lambda_1
      \approx \lambda_2$ (logarithmic, $50.000$ steps, subsampling
      $100$).}
    \label{fig_nearby}
  \end{center}
\end{figure}

Figure \ref{fig_nearby} shows the simulation results for nearby
eigenvalues $\lambda_1 \approx \lambda_2$.  Looking at the projection
error (right diagrams), both ``N2S'' and ``TwJ2S'' show slower
convergence than for evenly spaced eigenvalues (note the larger number
of simulation steps), but we see that ``N2S'' converges considerably
slower than ``TwJ2S'' which confirms the observation reported before
\cite[]{own_Moeller20a}. However, with increasing $\alpha$ in ``M2S'',
the time course of the projection error approaches that of
``TwJ2S''. Looking at the orthonormality error (left diagrams), we see
small values for exact back-projection, fast convergence for
approximated back-projection, and much slower convergence with no
back-projection. In the latter case, there is a tendency for faster
convergence with increasing $\alpha$ in ``M2S'', approaching ``TwJ2S''
for $\alpha=20$. Again, the projection error does not differ between
no back-projection and approximated back-projection, even though the
latter shows a noticeably faster reduction of the orthonormality
error.

All learning rules seem to approach a lower limit in both $e_o$ and
$e_p$ which can probably be explained by numerical effects.

%#############################################################################
%#############################################################################
%#############################################################################

\section{Discussion}\label{sec_discussion}
%===================

The simulations show a marked improvement of the convergence speed of
the modified learning rule ``M2S'' for increasing $\alpha$,
particularly if some principal eigenvalues are close to each
other. Nearby principal eigenvalues slow down both the ``N2S'' and the
``M2S'' rule, but the latter is affected more strongly for which we
can provide the explanation that the ``symmetry-breaking'' effect of
$\matD$ is reduced if the eigenvalue estimates on its diagonal are
close to each other \cite[]{own_Moeller20a}. Introducing the
additional terms in the modified objective function mitigates this
effect. However, we originally expected that the additional terms will
also modify or suppress the ``undesired'' fixed points, but our
analysis shows that all fixed points of ``N2S'' are also present in
``M2S''. We assume that the contributions by the different terms of
the additional sum cancel out in the fixed points which therefore
remain unchanged. Only the steepness of the landscape outside of the
fixed points is increased. Additional fixed points may be present in
``M2S'', but this was not analyzed here (particularly the different
constraint on $\matT$ may lead to additional fixed points).

It is always unfortunate if an additional parameter (in this case
$\alpha$) has to be introduced. We currently cannot provide a
universal guideline on how $\alpha$ has to be adjusted for different
eigenvalue spectra and dimensions. We observed that higher values of
$\alpha$ than the ones tested in the simulations sometimes lead to
divergence. We could imagine that learning rules can be designed where
$\alpha$ is suitably chosen depending on $\matW$. Note that also the
suitable range for the learning rate $\gamma$ is not clear. With exact
back-projection, it can be higher than with approximated
back-projection (since the approximation is based on the assumption of
small steps) or without back-projection, but suitable values may
depend on the eigenvalue spectrum and the dimensions.

We compare ``N2S'' and ``M2S'' with the learning rule ``TwJ2S'' where
we have a fixed diagonal weight-factor matrix $\matTheta$ with
distinct elements in the place of $\matD$ or $\matD'_\alpha$. The
influence of the choice of the elements of $\matTheta$ on the
convergence speed has to be studied. An absolute statement like `rule
``M2S'' performs better than ``TwJ2S'' for a certain $\alpha$' is
therefore debatable. The time course of the projection error of
``TwJ2S'' should therefore only be taken as a coarse reference.

The stability analysis in section \ref{sec_stability_mod} revealed
that the additional second term in the modified objective function
(\ref{eq_objfct_mod}) leads to a term depending only on the step
parameter $\matB$, see equations (\ref{eq_deltaJ2_final}) and
(\ref{eq_deltaJ2_final_mod}). With inverted sign, this term alone
would be sufficient to explain PCA behavior. One could therefore
assume that the additional term in objective function alone (with
inverted sign) could lead to a PCA rule. However, we have also shown
that the corresponding terms in the learning rule ``M2S'' are
characteristic for subspace behavior, thus the fixed-point structure
is completely different without the original first term.

We did not explore the difference between the ``short'' learning rules
studied here and the alternative of learning rules derived from the
``embedded'' metric on the Stiefel manifold.

%#############################################################################
%#############################################################################
%#############################################################################

\section{Conclusion}\label{sec_conclusion}
%===================

We introduced an additional term into the objective function which
improves the convergence speed of the corresponding learning rule,
particularly in the case of nearby principal eigenvalues. The modified
learning rule ``M2S'' is structurally similar to the original rule
``N2S'', with a different matrix $\matD'_\alpha$ in place of
$\matD$. Our analysis shows that the modified learning rule has all
fixed points of the original rule but may introduce new fixed points
(which was not studied further). Also the stability of the fixed
points shared with the original rule is unaffected by the
modification.

%#############################################################################
%#############################################################################
%#############################################################################

%\section*{Acknowledgments}
%%=========================
%\label{sec_acknowledgments}
%\addcontentsline{toc}{section}{\nameref{sec_acknowledgments}}

%#############################################################################
%#############################################################################
%#############################################################################

%\newpage
% homes <-> home
\trarxiv{%
  \bibliographystyle{/home/moeller/bst/plainnatsfnnm}
  \bibliography{/home/moeller/bib/nn17,/home/moeller/bib/own14}
}{%

}

%#############################################################################
%#############################################################################
%#############################################################################

\section*{Changes}
%=================
\label{sec_changes}
\addcontentsline{toc}{section}{\nameref{sec_changes}}

6 June 2020: Started report.\\
18 July 2020: Submission to arXiv.
%

%#############################################################################
%#############################################################################
%#############################################################################

\appendix

\section{Terms of Learning Rules Close to the Stiefel Manifold}
%==============================================================
\label{app_near_st}

In our previous work, we derived different learning rules, either from
a derivation in ``short'' or ``long'' form or from two different
metrics on the Stiefel manifold, canonical and embedded \exteq{section
  9}. The ``short'' rules coincide with the ``canonical'' rules, so we
have three groups: ``short'', ``long'', and ``embedded''.

In this report we focus on ``short'' learning rules. In simulations
(data not shown) comparing rules from the three groups for the
``original'' objective function (N2S, NL, NSE), the time course of the
projection error $e_p$ was not markedly different, regardless of the
back-projection method used. In the following we explore how the
different terms can be approximated if the learning rule operates in
the vicinity of the Stiefel manifold where $\matW^T\matW \approx
\matI_m$. We start from learning rule NL which contains all types of
terms known so far \exteq{476}:
\begin{align}
  \tau \matWdot
  &=
  5 \matC \matW \matD\nonumber\\
  &- \matW \matW^T \matC \matW \matD
  - \matW \matD \matW^T \matC \matW
  \nonumber\\
  &- \matC \matW \matD \matW^T \matW
  - \matC \matW \matD^*
  - \matC \matW \matW^T \matW \matD
\end{align}
where
\begin{align}
  \matD
  &=
  \Diag{j=1}{m}\{\vecw_j^T \matC \vecw_j\}
  = \dg\{\matW^T\matC\matW\}\\
  \matD^*
  &=
  \Diag{j=1}{m}\{\vecw_j^T \matC \matW \matW^T \vecw_j\}
  = \dg\{\matW^T\matC\matW\matW^T\matW\}.
\end{align}
Close to the Stiefel manifold, we have $\matD^* \approx
\dg\{\matW^T\matC\matW\} = \matD$. We can approximate the different
terms as
\begin{align}
  \tau \matWdot
  &=
  5 \matC \matW \matD\nonumber\\
  &- \matW \matW^T \matC \matW \matD
  - \matW \matD \matW^T \matC \matW
  \nonumber\\
  &- \matC \matW \matD
  - \matC \matW \matD
  - \matC \matW \matD
\end{align}
which leads to the rule called NSE \exteq{485}:
\begin{align}
  \tau \matWdot
  =
  2 \matC \matW \matD\nonumber
  - \matW \matW^T \matC \matW \matD
  - \matW \matD \matW^T \matC \matW.
\end{align}
There is no obvious approximation which leads from here to N2S, so we
assume that in the vicinity of the Stiefel manifold, there are
essentially just the two forms N2S and NSE, which correspond to the
gradient in the canonical or embedded metric \exteq{section 9.3},
respectively. Even these two rules show very similar behavior.

It is obvious that exact back-projection keeps $\matW$ on the Stiefel
manifold, and it is also clear that the approximated back-projection
almost achieves the same, at least for small learning rates $\gamma =
1 / \tau$. Why the rules return to the Stiefel manifold after each
learning step {\em without} back-projection remains to be explored.

\section{Additional Fixed Points of ``M2S''}
%===========================================
\label{app_add_fp}

\begin{figure}[t]
  \begin{center}
    % space required to avoid problem with bounding box
    \includegraphics[width=10cm]{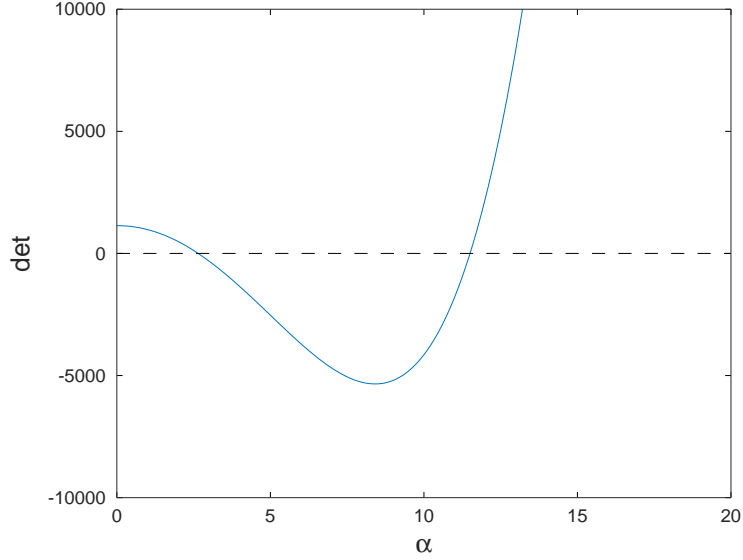}\\[10mm]
    \caption{Determinant $\det\{\matD'_{\alpha}\}$ over $\alpha$ for a
      random, semi-orthogonal $\matW$ of size $10 \times 4$ and
      eigenvalues descending from $10.0$ to $1.0$ in steps of $1.0$.}
    \label{fig_det_Dp_alpha}
  \end{center}
\end{figure}

 We analyze whether the second factor in equation
 (\ref{eq_M2S_constraint_T}) can become singular:
\begin{align}
  \matDnull'_\alpha
  &=
  (1+\alpha) \dg\{\matWnull^T\matC\matWnull\}
  -
  \alpha \matWnull^T\matC\matWnull\\
  &=
   (1+\alpha) \dg\{\matAnull^T\matV^T\matC\matV\matAnull\}
  -
  \alpha \matAnull^T\matV^T\matC\matV\matAnull\\
  &=
   (1+\alpha) \dg\{\matAnull^T\matLambda\matAnull\}
  -
  \alpha \matAnull^T\matLambda\matAnull.
\end{align}
In a simulation, we generate a random semi-orthogonal $\matA$ of size
$n \times m$ (with $n=10$, $m=4$) and use eigenvalues $\{n, n-1,
\ldots, 1\}$ to form $\matLambda$. We vary $\alpha$ and plot
$\det\{\matD'_{\alpha}\}$ in steps of $0.1$ from $0.0$ to $20.0$ in
figure \ref{fig_det_Dp_alpha}. We often see two zero-crossings as
shown in the figure, but curves with other shapes appear as well, depending on the random initialization of $\matA$.
  
\section{Fixed-Point Constraints of ``M2S''}
%===========================================
\label{app_constraints_S_T}

We describe two attempts at deriving the constraints on matrices
$\matS$ and $\matT$ which lead to the same result.

\subsection{Attempt 1}
%---------------------

The first attempt starts from (\ref{eq_M2S_form1}):
\begin{align}
  \nonumber
  \matNull
  &=
  (1 + \alpha)
  (\matC\matWnull\matDnull
  -\matWnull\matDnull\matWnull^T\matC\matWnull)\\
  &-
  \alpha
  (\matC\matWnull
  -\matWnull\matWnull^T\matC\matWnull)
  (\matWnull^T\matC\matWnull)\\[5mm]
  %-----
  \nonumber
  \matNull
  &=
  (1 + \alpha)
  (\matC\matV\matAnull\matDnull
  -\matV\matAnull\matDnull\matAnull^T\matV^T\matC
  \matV\matAnull)\\
  &-
  \alpha
  (\matC\matV\matAnull
  -\matV\matAnull\matAnull^T\matV^T\matC\matV\matAnull)
  (\matAnull^T\matV^T\matC\matV\matAnull)\\[5mm]
  %-----
  \nonumber
  \matNull
  &=
  (1 + \alpha)
  (\matC\matV\matAnull\matDnull
  -\matV\matAnull\matDnull\matAnull^T\matLambda\matAnull)\\
  &-
  \alpha
  (\matC\matV\matAnull
  -\matV\matAnull\matAnull^T\matLambda\matAnull)
  (\matAnull^T\matLambda\matAnull)\\[5mm]
  %-----
  \nonumber
  \matNull
  &=
  (1 + \alpha)
  (\matV^T\matC\matV\matAnull\matDnull
  -\matV^T\matV\matAnull\matDnull\matAnull^T\matLambda\matAnull)\\
  &-
  \alpha
  (\matV^T\matC\matV\matAnull
  -\matV^T\matV\matAnull\matAnull^T\matLambda\matAnull)
  (\matAnull^T\matLambda\matAnull)\\[5mm]
  %-----
  \nonumber
  \matNull
  &=
  (1 + \alpha)
  (\matLambda\matAnull\matDnull
  -\matAnull\matDnull\matAnull^T\matLambda\matAnull)\\
  &-
  \alpha
  (\matLambda\matAnull
  -\matAnull\matAnull^T\matLambda\matAnull)
  (\matAnull^T\matLambda\matAnull)\\[5mm]
  %-----
  \nonumber
  \matNull
  &=
  (1 + \alpha)
  \left[\matLambda\matQ\pmat{\matI_m\\ \matNull}\matDnull
  -\matQ\pmat{\matI_m\\ \matNull}\matDnull\pmat{\matI_m & \matNull^T}\matQ^T
  \matLambda\matQ\pmat{\matI_m\\ \matNull}\right]\\
  &-
  \alpha
  \left[\matLambda\matQ\pmat{\matI_m\\ \matNull}
  -\matQ\pmat{\matI_m\\ \matNull}\pmat{\matI_m & \matNull^T}\matQ^T
  \matLambda\matQ\pmat{\matI_m\\ \matNull}\right]
  \left[\pmat{\matI_m & \matNull^T}\matQ^T
    \matLambda
    \matQ\pmat{\matI_m\\ \matNull}\right]\\[5mm]
  %-----
  \nonumber
  \matNull
  &=
  (1 + \alpha)
  \left[\matQ^T\matLambda\matQ\pmat{\matI_m\\ \matNull}\matDnull
    -\matQ^T\matQ\pmat{\matI_m\\ \matNull}\matDnull\pmat{\matI_m & \matNull^T}
    \matQ^T\matLambda\matQ\pmat{\matI_m\\ \matNull}\right]\\
  \nonumber
  &-
  \alpha
  \left[\matQ^T\matLambda\matQ\pmat{\matI_m\\ \matNull}
    -\matQ^T\matQ\pmat{\matI_m\\ \matNull}\pmat{\matI_m & \matNull^T}
    \matQ^T\matLambda\matQ
    \pmat{\matI_m\\ \matNull}\right]\\
  &\cdot
  \left[\pmat{\matI_m & \matNull^T}
    \matQ^T\matLambda\matQ
    \pmat{\matI_m\\ \matNull}\right]\\[5mm]
  %-----  
  \nonumber
  \matNull
  &=
  (1 + \alpha)
  \left[\matM\pmat{\matI_m\\ \matNull}\matDnull
    -\pmat{\matI_m\\ \matNull}\matDnull\pmat{\matI_m & \matNull^T}
    \matM\pmat{\matI_m\\ \matNull}\right]\\
  &-
  \alpha
  \left[\matM\pmat{\matI_m\\ \matNull}
    -\pmat{\matI_m\\ \matNull}\pmat{\matI_m & \matNull^T}
    \matM
    \pmat{\matI_m\\ \matNull}\right]
  \left[\pmat{\matI_m & \matNull^T}
    \matM\pmat{\matI_m\\ \matNull}\right]\\[5mm]
  %-----  
  \nonumber
  \matNull
  &=
  (1 + \alpha)
  \left[\pmat{\matS\\ \matT}\matDnull
    -\pmat{\matI_m\\ \matNull}\matDnull
    \matS\right]\\
  &-
  \alpha
  \left[\pmat{\matS\\ \matT}
    -\pmat{\matI_m\\ \matNull}\matS\right]
  \matS\\[5mm]
  %-----  
  \nonumber
  \matNull
  &=
  (1 + \alpha)
  \left[\pmat{\matS\matDnull\\ \matT\matDnull}
    -\pmat{\matDnull\matS\\ \matNull}\right]\\
  &-
  \alpha
  \left[\pmat{\matS^2\\ \matT\matS}
    -\pmat{\matS^2\\ \matNull}\right]\\[5mm]
  %-----
  \label{eq_M2S_constraint_1_matrix}
  \matNull
  &=
  \pmat{
    (1+\alpha)[\matS\matDnull-\matDnull\matS]\\
    \matT[(1+\alpha)\matDnull-\alpha\matS]}.
\end{align}
The upper part of equation (\ref{eq_M2S_constraint_1_matrix}) gives
\begin{align}
  \label{eq_M2S_constraint_1}
  \matS\matDnull
  &=
  \matDnull\matS.
\end{align}
Equation (\ref{eq_M2S_constraint_1}) coincides with the constraint
\exteq{247} derived for the fixed points of learning rule ``N2S''.

The lower part of equation (\ref{eq_M2S_constraint_1_matrix}) gives
\begin{align}
\matT[(1+\alpha)\matDnull-\alpha\matS] = \matNull
\end{align}
which differs from the equation $\matT\matDnull = \matNull$ derived
for ``N2S''.

\subsection{Attempt 2}
%---------------------

The second attempt starts from (\ref{eq_M2S_form2}) and proceeds in
the same way as in section 7.8 of our previous work
\cite[]{own_Moeller20a}, from equation \exteq{239} onward, except with
$\matDnull'_\alpha$ instead of $\matDnull$:
\begin{align}
  \matNull
  &= \matC \matWnull \matDnull'_\alpha
  - \matWnull \matDnull'_\alpha \matWnull^T \matC \matWnull\\
  \matNull
  &=
  \matC \matV \matAnull \matDnull'_\alpha
  - \matV \matAnull \matDnull'_\alpha \matAnull^T
  \matV^T \matC \matV \matAnull\\
   \matNull
  &=
  \matV^T \matC \matV \matAnull \matDnull'_\alpha
  - \matV^T\matV \matAnull \matDnull'_\alpha \matAnull^T
  \matV^T \matC \matV \matAnull\\
  \label{eq_M2S_intermediate}
  \matNull
  &=
  \matLambda \matAnull \matDnull'_\alpha
  - \matAnull \matDnull'_\alpha \matAnull^T \matLambda \matAnull\\
  \matNull
  &=
  \matLambda \matQ \pmat{\matI_m\\ \matNull} \matDnull'_\alpha
  - \matQ \pmat{\matI_m\\ \matNull}
  \matDnull'_\alpha
  \pmat{\matI_m & \matNull^T}
  \matQ^T \matLambda \matQ \pmat{\matI_m\\ \matNull}\\
  \matNull
  &=
  \matQ^T \matLambda \matQ \pmat{\matI_m\\ \matNull} \matDnull'_\alpha
  - \pmat{\matI_m\\ \matNull} \matDnull'_\alpha \pmat{\matI_m & \matNull^T}
  \matQ^T \matLambda \matQ \pmat{\matI_m\\ \matNull}\\
  \matNull
  &=
  \pmat{\matS & \matT^T\\\matT & \matU} \pmat{\matI_m\\ \matNull}
  \matDnull'_\alpha
  - \pmat{\matI_m\\ \matNull} \matDnull'_\alpha \pmat{\matI_m & \matNull^T}
  \pmat{\matS & \matT^T\\\matT & \matU} \pmat{\matI_m\\ \matNull}\\
  \label{eq_M2S_constraint_2}
  \pmat{\matNull\\ \matNull}
  &=
  \pmat{\matS \matDnull'_\alpha \\ \matT \matDnull'_\alpha}
  - \pmat{\matDnull'_\alpha \matS\\ \matNull}.
\end{align}
To analyze the constraint on $\matS$ in the upper equation of
(\ref{eq_M2S_constraint_2}), we look at
\begin{align}
\matDnull'_\alpha = (1 + \alpha) \matDnull - \alpha \matWnull^T\matC\matWnull
\end{align}
and see that
\begin{align}
  \matWnull^T\matC\matWnull
  &=
  \matAnull^T\matV^T\matC\matV\matAnull\\
  &=
  \matAnull^T\matLambda\matAnull\\
  &=
  \pmat{\matI_m & \matNull^T}
  \underbrace{\matQ^T\matLambda\matQ}_{\matM}
  \pmat{\matI_m\\ \matNull}\\
  &=
  \pmat{\matI_m & \matNull^T}
  \pmat{\matS & \matT^T\\ \matT & \matU}
  \pmat{\matI_m\\ \matNull}\\
  &=
  \matS.
\end{align}
Note that we also have $\matDnull = \dg\{\matWnull^T\matC\matWnull\} =
\dg\{\matS\}$.

We can therefore write $\matDnull'_\alpha$ as
\begin{align}
\matDnull'_\alpha = (1 + \alpha) \matDnull - \alpha \matS.
\end{align}
We proceed with the upper equation of (\ref{eq_M2S_constraint_2}):
\begin{align}
  \matS\matDnull'_\alpha
  &=
  \matDnull'_\alpha \matS\\
  \matS\left[(1+\alpha)\matDnull-\alpha\matS\right]
  &=
  \left[(1+\alpha)\matDnull-\alpha\matS\right]\matS\\
  (1+\alpha)\matS\matDnull - \alpha\matS^2
  &=
  (1+\alpha)\matDnull\matS - \alpha\matS^2\\
  \matS\matDnull
  &=
  \matDnull\matS.
\end{align}
This constraint is the same as \exteq{247} which was derived for the
fixed points of the ``N2S'' learning rule.

We now look at the lower equation of (\ref{eq_M2S_constraint_2}):
\begin{align}
  \matT \matDnull'_\alpha
  &=
  \matNull\\
  \matT \left[(1+\alpha)\matDnull-\alpha\matS\right]
  &=
  \matNull.
\end{align}
The corresponding equation for ``N2S'' was $\matT\matDnull=\matNull$.

%#############################################################################
% Lemmata
%#############################################################################

%\section{Lemmata}
%================

\end{document}